\documentclass{article}

\usepackage{style/arxiv}

\usepackage[utf8]{inputenc} 
\usepackage[T1]{fontenc}    
\usepackage{hyperref}       
\usepackage{url}            
\usepackage{booktabs}       
\usepackage{amsfonts}       
\usepackage{nicefrac}       
\usepackage{microtype}      
\usepackage{lipsum}
\usepackage{graphicx}
\usepackage{tabularx}
\usepackage{array}
\usepackage{multirow}
\usepackage{makecell}
\usepackage{siunitx}
\usepackage[table]{xcolor}
\usepackage{colortbl}
\usepackage{footnote}
\usepackage{ragged2e}
\usepackage{algorithm}
\usepackage{algpseudocode}
\usepackage{amssymb}
\usepackage{textcomp}
\usepackage{float}
\usepackage{enumitem}
\usepackage{cleveref}
\usepackage[framemethod=TikZ]{mdframed}

\definecolor{llama}{RGB}{225,235,250}       
\definecolor{gemma}{RGB}{235,225,245}       
\definecolor{mistral}{RGB}{255,235,215}     
\definecolor{encoder}{RGB}{245,245,245}     
\definecolor{yes}{RGB}{220, 245, 225}       
\definecolor{no}{RGB}{245, 220, 220}        
\definecolor{promptbg}{RGB}{248,248,248}    
\definecolor{promptbg1}{HTML}{ffd6a5}
\definecolor{promptbg2}{HTML}{a9def9}
\definecolor{promptbg3}{HTML}{fcf6bd}

\newcommand{\deltacell}[1]{\makecell[c]{#1}}
\newcolumntype{Y}{>{\centering\arraybackslash}X}

\graphicspath{ {./images/} }

\title{xList-Hate: A Checklist-Based Framework for Interpretable and Generalizable Hate Speech Detection}

\author{
 Adrián Girón \\
  Universidad Politécnica de Madrid\\
  Madrid, Spain \\
  \texttt{adrian.giron@upm.es} \\
   \And
 Pablo Miralles \\
  Universidad Politécnica de Madrid\\
  Madrid, Spain \\
  \texttt{pablo.miralles@upm.es} \\
   \And  
 Javier Huertas-Tato \\
  Universidad Politécnica de Madrid\\
  Madrid, Spain \\
  \texttt{javier.huertas.tato@upm.es} \\
   \And
 Sergio D'Antonio \\
  Universidad Politécnica de Madrid\\
  Madrid, Spain \\
  \texttt{sergio.dantonio@upm.es} \\
   \And
 David Camacho \\
  Universidad Politécnica de Madrid\\
  Madrid, Spain \\
  \texttt{david.camacho@upm.es} \\
}

\begin{document}
\maketitle
\begin{abstract}
Hate speech detection is commonly framed as a direct binary classification problem despite  being a composite concept defined through multiple interacting factors that vary across legal frameworks, platform policies, and annotation guidelines. As a result, supervised models often overfit dataset-specific definitions and exhibit limited robustness under \textit{domain shift} and \textit{annotation noise}.

We introduce \texttt{xList-Hate}, a diagnostic framework that decomposes hate speech detection into a checklist of explicit, concept-level questions grounded in widely shared normative criteria. Each question is independently answered by a large language model (LLM), producing a binary diagnostic representation that captures hateful content features without directly predicting the final label. These diagnostic signals are then aggregated by a lightweight, \textit{fully interpretable} decision tree, yielding transparent and auditable predictions.

We evaluate it across multiple hate speech benchmarks and model families, comparing it against zero-shot LLM classification and in-domain supervised fine-tuning. While supervised methods typically maximize in-domain performance, we consistently improves cross-dataset robustness and relative performance under domain shift. In addition, qualitative analysis of disagreement cases provides evidence that the framework can be less sensitive to certain forms of annotation inconsistency and contextual ambiguity. Crucially, the approach enables fine-grained interpretability through explicit decision paths and factor-level analysis.

Our results suggest that reframing hate speech detection as a diagnostic reasoning task, rather than a monolithic classification problem, provides a robust, explainable, and extensible alternative for content moderation.
\end{abstract}


\begin{mdframed}[
  linewidth=1.2pt,
  roundcorner=3pt,
  backgroundcolor=promptbg,
]
\begin{center}
    \textbf{Content Warning}: This paper analyzes actual hate speech samples in a research context.
\end{center}
\end{mdframed}

\section{Introduction}
\label{sec:introduction}

Hate speech has become a central challenge in online communication, particularly in the context of social media platforms where large volumes of user-generated content are produced and disseminated at scale~\cite{ExposureHateOnline2025, HateSpeechEpidemic2020}. The automatic detection of hate speech is critical for content moderation, legal compliance, and the protection of vulnerable groups~\cite{PrevalencePsychologicalEffects2019}. However, despite extensive research efforts, hate speech detection remains a difficult and unresolved problem.

One of the core difficulties lies in the lack of a single, universally accepted definition of hate speech. Legal frameworks, platform policies, and academic taxonomies vary significantly in how they define and operationalize the concept, depending on legal standards, cultural norms, and contextual considerations. As a result, hate speech is not a monolithic phenomenon but rather a collection of related conceptual factors, such as targeting of protected groups, derogatory or dehumanizing language, incitement to discrimination, and calls for violence. This definitional variability introduces substantial ambiguity into both dataset annotation and model evaluation, making robust automatic detection particularly challenging.

Recent advances in large language models (LLMs) have led to their widespread use for hate speech detection~\cite{piotCanLLMsEvaluate2025}. Zero-shot and few-shot prompting approaches offer an appealing level of flexibility, but in practice they often yield inconsistent predictions and only moderate performance. More importantly, their decisions are opaque and difficult to justify, which is problematic in a sensitive domain such as hate speech moderation.

Supervised fine-tuning of language models typically achieves stronger in-domain performance, but this improvement often comes at the cost of poor generalization across datasets~\cite{yinGeneralisableHateSpeech2021}. This limitation is exacerbated by the scarcity of high-quality, diverse training data and by the inherent subjectivity of hate speech annotations, which can encode dataset-specific biases and annotation artifacts~\cite{LargeLanguageModel2025}. As a consequence, current approaches struggle to simultaneously achieve strong transferability, robustness, and interpretability.

In this work, we propose an alternative approach to hate speech detection based on a diagnostic perspective. Instead of asking a model to directly predict whether a text constitutes hate speech, we decompose the task into a set of ten binary questions, each capturing a distinct conceptual factor commonly associated with hate speech. These questions are designed to be as objective and self-contained as possible and are grounded in established legal definitions, platform policies, and academic frameworks. For each input text, a large language model answers the ten questions independently, producing a binary vector that represents a structured diagnostic profile of the content.

\begin{figure}
    \centering
    \includegraphics[width=0.9\linewidth]{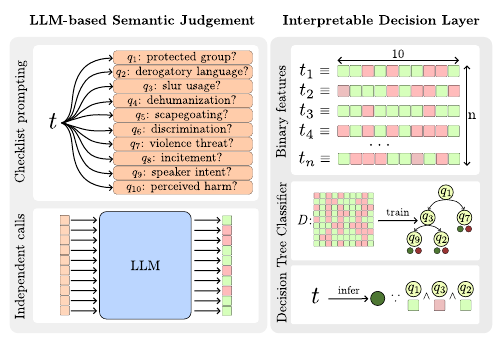}
    \caption{Overview of the proposed checklist-based hate speech detection framework.
    Each input text $t$ is evaluated through a fixed set of ten diagnostic questions, answered independently by an LLM.
    The resulting feature vector is then passed to an interpretable decision layer, implemented as a lightweight decision tree, which performs the final classification.}

    \label{fig:framework}
\end{figure}

The final classification decision is then delegated to a lightweight and fully interpretable model, specifically a decision tree, trained on these binary representations. This design intentionally separates semantic judgment from classification: the expressive capacity of the LLM is used to evaluate well-defined conceptual criteria, while the final decision logic remains transparent and easily interpretable. As a result, the proposed pipeline is modular, low-resource at the training stage, and inherently explainable. An overview of our framework is shown in~\Cref{fig:framework}.

We evaluate our method across multiple hate speech datasets and compare it against zero-shot LLM prompting and supervised fine-tuning baselines. While fine-tuned models often achieve higher performance in-domain, our approach consistently generalizes better out-of-domain. In addition, the use of decision trees enables the analysis of which conceptual factors contribute to a decision, making our approach more interpretable. Further, the small and interpretable feature space can be used to compare differences in distribution and biases across datasets.


In summary, our contributions are as follows:
\begin{itemize}
    \item We introduce xList-Hate, a question-based diagnostic framework for hate speech detection that replaces abstract hate/neutral predictions with a set of independent, conceptually grounded binary judgements derived from legal, policy, and academic definitions.
    \item The xList-Hate framework enables transparent and auditable decision-making by grounding final predictions in explicit diagnostic factors. By training a lightweight and interpretable supervised model (such as decision trees) on top of these factors, we allow full traceability from individual checklist answers to the final classification and facilitate fine-grained bias and error analysis.
    \item We empirically compare this framework with supervised fine-tuning. In-domain performance is lower, although the gap diminishes as model size increases. On the other hand, xList-Hate is more robust and stable in performance across distribution shifts. Our framework is even competitive or superior in out-of-domain absolute performance in several training sets and backbone models.
    \item We provide theoretical arguments and dataset samples showing that this approach yields models that are more robust to annotation noise and dataset-specific labeling criteria, matching our experiments in cross-dataset generalization and robustness.
\end{itemize}
\section{Related Work}
\label{sec:related_work}



\subsection{Hate speech definitions and annotation practices}

A persistent challenge in hate speech research is that hate speech is not a single, universally operationalized construct: legal, institutional, and scholarly definitions diverge substantially in both scope and enforcement thresholds. Legal-oriented treatments often foreground \emph{incitement}, especially incitement to violence or criminal acts, placing less emphasis on milder but still harmful forms such as discriminatory or exclusionary rhetoric \cite{DefiningHateSpeech2020}. Broader reviews similarly highlight strong cross-national variation in hate speech legislation and enforcement, where commitments to freedom of expression shape what is considered sanctionable or even definitional; they also note that societal shocks (e.g., terrorist attacks) can catalyze online hate speech production and diffusion, and that some definitions insufficiently incorporate victim-centered perception and harm \cite{HateSpeechReview2018}.

This definitional instability propagates directly into empirical research. A large-scale mapping of the scientific literature finds substantial variation in which identities are treated as ``protected'' and which behaviors count as hateful, alongside systematic skew in detection research toward racial and religious targets relative to other protected attributes such as gender, sexuality, or disability \cite{MappingScientificKnowledge2024}. The same work reports geographic asymmetries, with hate crime origins appearing more concentrated in English-speaking contexts (notably the United States), while European contexts exhibit comparatively stronger regulatory and legislative activity. Crucially, it identifies a recurring disconnect between \emph{conceptual} work on hate speech definitions and \emph{methodological} work that focuses almost exclusively on maximizing performance metrics on labeled datasets without a stable conceptual grounding \cite{MappingScientificKnowledge2024}.

This heterogeneity is also visible in applied measurement instruments and domain-specific reviews. A systematic review focusing on youth-facing contexts finds that both definitions and assessment tools vary widely, that exposure to hate speech is more frequent than reported victimization or perpetration, and that hate speech frequently overlaps with (and is sometimes conflated with) bullying and harassment \cite{SystematicReviewHate2023}. Complementing this, bibliometric evidence over three decades documents a pronounced shift in the last 10-15 years from theoretical frameworks toward operational detection systems, largely driven by the rise of social media, often at the expense of interpretability and conceptual analysis \cite{ThirtyYearsResearch2021}. This shift is particularly problematic given well-documented annotation variability across annotator backgrounds, cultures, and normative assumptions.

Recent work explicitly connects this conceptual-annotation mismatch to inflated performance claims in hate speech detection. Cross-dataset evaluations show substantial performance drops when models are trained and evaluated out-of-domain, revealing that many reported state-of-the-art gains stem from dataset-specific artifacts, token-label correlations, and uneven sampling of hate speech phenomena rather than robust conceptual learning \cite{arangoHateSpeechDetection2022}. Related analyses demonstrate that supervised Transformer-based models often over-rely on lexical ``keyword'' cues, allocating disproportionate attention to hate-associated tokens even when they are not decisive for the underlying concept \cite{delapenasarracenSystematicKeywordBias2023}. In line with these findings, work explicitly aimed at generalisable hate speech detection emphasizes cross-dataset testing as a necessary evaluation protocol and attributes poor transfer to both technical limitations of supervised NLP pipelines and dataset-to-dataset variation in labeling criteria and definitions \cite{yinGeneralisableHateSpeech2021}.

Finally, recent studies have begun to examine annotation bias more directly in the context of both humans and LLMs. Evidence suggests that while both human annotators and LLMs exhibit systematic biases, these biases differ qualitatively: annotator demographics and group membership shape labeling patterns, whereas LLM biases emerge from training data distributions and alignment objectives \cite{giorgiHumanLLMBiases2025}. Survey-style reviews conclude that although LLMs improve contextual handling in many scenarios, bias mitigation, interpretability, multilingual robustness, and transparency remain open challenges \cite{HateSpeechDetection2025}. Other work shows that even within zero-shot LLM settings, performance is sensitive to how hate speech is defined: constructing modular taxonomies and systematically varying definitional components can substantially alter predictions, underscoring that definitions themselves function as modeling choices rather than neutral background assumptions \cite{ModularTaxonomyHate2025}.

Taken together, prior work establishes that (i) hate speech definitions vary widely across legal, cultural, and institutional contexts; (ii) datasets inherit this variability through heterogeneous guidelines and annotation subjectivity; and (iii) supervised detectors frequently learn dataset-specific shortcuts that undermine cross-dataset robustness. These findings motivate approaches that preserve explicit conceptual structure while remaining adaptable to shifting definitions and annotation regimes.

\subsection{Supervised hate speech classification}

The dominant paradigm in hate speech detection has historically relied on supervised learning, ranging from classical machine learning pipelines to deep neural architectures and, more recently, fine-tuned large language models. These systems typically frame hate speech detection as a binary or multi-class classification problem trained on annotated datasets, often reporting strong in-domain performance. However, a growing body of work has highlighted fundamental limitations of this paradigm, particularly regarding generalization, bias, and interpretability.

A central critique concerns the reliability of reported state-of-the-art results. Large-scale re-evaluations show that many headline gains are inflated due to experimental practices such as training and testing in-domain rather than under realistic cross-dataset conditions \cite{arangoHateSpeechDetection2022}. When evaluated out-of-domain, these models often suffer dramatic performance drops, revealing substantial overfitting to dataset-specific artifacts, including lexical correlations, annotation conventions, and uneven sampling of hate speech subtypes. Similar conclusions are drawn in work emphasizing generalisable hate speech detection, which attributes poor transfer to both technical limitations of supervised NLP models and variability in dataset definitions and labeling criteria \cite{yinGeneralisableHateSpeech2021}.

Despite these limitations, supervised approaches remain prevalent. Recent studies explore increasingly complex architectures, including multimodal systems that combine CNNs, RNNs, and Transformer components for joint text, image hate speech detection, typically evaluated exclusively in-domain \cite{ComprehensiveFrameworkMultimodal2025}. Other work fine-tunes LLMs directly for hate speech classification, including language-specific settings such as Bengali, again focusing on in-domain evaluation and reporting interpretability via local explanation tools such as LIME \cite{DeepHateDetectComparativeStudy2025}. Similar patterns appear in Arabic hate speech detection, where self-supervised pretraining is followed by supervised classification and post-hoc explanation using LIME \cite{InterpretableArabicHate2025}.

Interpretability in supervised systems is most commonly addressed through post-hoc attribution techniques such as LIME or SHAP. While these methods can highlight influential tokens, they remain inherently tied to surface-level textual features. Empirical analyses show that supervised Transformer models often concentrate attention on a narrow set of hate-associated keywords, reinforcing shortcut learning and domain-specific bias \cite{delapenasarracenSystematicKeywordBias2023}. This limitation extends to approaches that incorporate LLM-generated rationales: although explanations are produced explicitly, downstream encoder models are still trained and evaluated in-domain, and interpretability claims remain grounded in token-level attributions rather than explicit conceptual reasoning \cite{LargeLanguageModel2025,UnderstandingHateSpeech2024}.

Overall, the supervised literature demonstrates that high in-domain accuracy can be achieved across languages, modalities, and architectures \cite{SaferOnlineCommunities2024,ComprehensiveFrameworkMultimodal2025}. However, these gains often come at the expense of robustness and conceptual transparency. Models are sensitive to annotation noise, dataset-specific definitions, and lexical correlations, and require costly retraining when domains or labeling criteria change. Without explicit mechanisms to disentangle conceptual dimensions of hate speech from surface cues, supervised detectors struggle to generalize and offer limited insight into why a particular decision was made.

\subsection{Inference-based hate speech classification}

The emergence of large language models has enabled a shift toward inference-based hate speech classification, where predictions are produced directly at inference time via prompting rather than learned from labeled data. Zero-shot and few-shot approaches have demonstrated promising robustness across heterogeneous datasets, particularly when samples originate from different regions and annotation cultures \cite{EvaluationHateSpeech2025}. This suggests reduced dependence on dataset-specific annotation artifacts compared to supervised models.

At the same time, multiple studies reveal substantial inconsistency in LLM-based moderation. Different LLMs often disagree markedly when evaluating the same content, indicating that inference-based systems do not implement a single, stable notion of hate speech \cite{giorgiHumanLLMBiases2025}. Comparative analyses of human and LLM annotators further show that while both exhibit systematic biases, these biases differ qualitatively: human judgments are shaped by annotator background and group membership, whereas LLM biases reflect training data and alignment objectives \cite{faschingModelDependentModerationInconsistencies2025}. Studies of inter-annotator agreement conclude that, despite their fluency, LLMs remain less reliable than humans when used as standalone labeling agents \cite{piotCanLLMsEvaluate2025}.

Hybrid approaches attempt to combine inference and supervision by prompting LLMs to generate rationales, then training encoder-based classifiers on raw text, rationales, or both. Examples include fine-tuning BERT-style models on LLM-generated explanations and claiming interpretability via SHAP or LIME \cite{LargeLanguageModel2025,UnderstandingHateSpeech2024}. Other work concatenates representations from raw text and rationales to perform supervised classification \cite{TARGELargeLanguage2025}. However, these pipelines remain fundamentally supervised, are evaluated in-domain, and risk inheriting both annotation bias and shortcut learning. Moreover, interpretability is indirect: explanations are mediated through token attributions, and causal links between rationales and predictions are diluted by supervised training.

A complementary line of work explores prompt-based definitional decomposition. Modular taxonomies derived from legal and policy sources show that zero-shot LLM performance is sensitive to which definitional components are included or excluded, highlighting that definitions themselves act as implicit modeling choices \cite{ModularTaxonomyHate2025}.

Overall, inference-based hate speech detection reduces reliance on labeled data and improves contextual sensitivity, but prior work highlights unresolved challenges: moderation criteria vary across models, biases persist, and most explanations remain post-hoc or token-centric \cite{HateSpeechDetection2025}. These limitations motivate frameworks that make the decision process explicit, modular, and aligned with well-defined conceptual questions.

\subsection{Explainable hate speech detection}

Explainability has become increasingly important in hate speech detection due to legal, ethical, and operational requirements. Most existing work, however, treats explainability as a post-hoc diagnostic layered on top of supervised classifiers. Recent surveys place the state of the art around encoder-based models combined with attribution methods such as SHAP or LIME \cite{DecodingFakeNews2025}. While these techniques provide insights into influential tokens, they offer limited access to higher-level conceptual reasoning.

Several supervised systems explicitly claim interpretability through attribution methods. Architectures based on BiRNNs, BiLSTMs, or BERT-like models apply SHAP or LIME to highlight salient words, typically under in-domain evaluation \cite{SaferOnlineCommunities2024,InterpretableArabicHate2025,DeepHateDetectComparativeStudy2025}. These explanations remain inherently lexical, revealing correlations rather than decision logic grounded in definitions of hate speech.

More recent work incorporates LLMs to generate textual rationales as intermediate explanations. In these pipelines, LLMs are prompted to justify why an input is hateful or not, and supervised classifiers are trained on these explanations or their representations \cite{UnderstandingHateSpeech2024,TARGELargeLanguage2025}. Interpretability is then inferred from either the rationales themselves or token attributions over the downstream model. However, the final decision is still produced by a supervised system whose internal reasoning may not preserve the structure or intent of the rationales, and evaluations remain in-domain, limiting claims of robustness.

Overall, existing approaches predominantly treat explanation as an auxiliary artifact rather than an intrinsic property of the classification process. Token-level attributions and post-hoc rationales do not expose how abstract factors such as endorsement, dehumanization, discrimination, or incitement interact to produce a decision. This gap motivates approaches that embed interpretability directly into the model architecture, where predictions arise from explicit, concept-level judgments that can be traced, inspected, and analyzed independently of dataset-specific supervision.

\section{Method}
\label{sec:method}

\subsection{Checklist: a diagnostic, factorized view of hate speech}
\label{sec:checklist}

Hate speech is not a basic concept but a composite one, defined through a set of recurring conditions that appear across legal standards, platform policies, and normative frameworks.
While these systems differ in enforcement thresholds and scope, they converge on a small number of conceptual dimensions that determine whether an expression qualifies as hate speech.
Rather than adopting a single operational definition, we derive our checklist by explicitly decomposing hate speech into these minimal, widely shared factors. These factors are transformed into questions with a binary answer, which are shown in~\Cref{tab:checklist_prompts}). We categorize and choose the following factors and questions:

\begin{table*}[!ht]
\centering
\footnotesize
\setlength{\tabcolsep}{6pt}
\renewcommand{\arraystretch}{1.25}
\begin{tabular}{
p{0.14\textwidth}
p{0.14\textwidth}
p{0.14\textwidth}
p{0.14\textwidth}
p{0.14\textwidth}
p{0.14\textwidth}
}
\toprule
\textbf{Source} &
\textbf{Protected Target} &
\textbf{Derogatory / Attacking Content} &
\textbf{Dehumanization / stereotyping / scapegoating} &
\textbf{Discrimination / exclusion} &
\textbf{Violence / incitement \& context qualifiers} \\
\midrule

\multicolumn{6}{l}{\textbf{International organizations and civil-society frameworks}} \\
\addlinespace[2pt]

United Nations~\cite{un_hate_speech} &
``...targeting a group or an individual based on inherent characteristics...'' &
``...offensive discourse...'' &
\begin{center}-\end{center} &
``...uses pejorative or discriminatory language...'' &
``...incitement to... violence...'' \\

Council of Europe (ECRI)~\cite{coe_hate_speech} &
``...against a person or group of persons for a variety of reasons...'' &
``...advocate, incite, promote or justify hatred...'' &
\begin{center}-\end{center} &
``...advocate, incite, promote... discrimination...'' &
``...advocate, incite, promote... violence...'' \\

UNESCO~\cite{unesco_hate_speech} &
``...a person or a group on the basis of who they are...'' &
``...that attacks or uses pejorative or discriminatory language...'' &
``...scapegoating, stereotyping, stigmatization...'' &
``...constitute incitement to... discrimination...'' &
``...constitute incitement to violence, hostility...'' \\

Rights for Peace~\cite{rightsforpeace_hate_speech} &
``incites discrimination, hostility or violence'' &
``...incites... hostility...'' &
\begin{center}-\end{center} &
``...incites discrimination...'' &
``...incites... violence...'' \\

ALA~\cite{ala_hate_speech} &
``...on the basis of race, religion, skin color, sexual identity...'' &
``...intend to vilify, humiliate...'' &
\begin{center}-\end{center} &
``...deprivation of civil rights...'' &
``...incite hatred...conspiracy to commit these crimes...'' \\

\addlinespace[4pt]
\multicolumn{6}{l}{\textbf{Social media platform policies}} \\
\addlinespace[2pt]

Meta (Facebook \& Instagram)~\cite{meta_hateful_conduct} &
``...against people... on the basis of what we call protected characteristics'' &
``...direct attacks..'' &
``...dehumanising speech... harmful stereotypes...'' &
``...these words are tied to historical discrimination...'' &
``...these words are tied to... violence...'' \\

YouTube~\cite{youtube_hate_speech} &
``...against individuals or groups based on...a protected group status...'' &
``...promotes... hatred...'' &
``...(people with protected group status) are like animals....'' &
``They got what they deserved (referring to people with protected group status)...'' &
``...promotes... violence...'' \\

X (Twitter)~\cite{x_hateful_conduct} &
``...on the basis of religion, caste, age, disability, disease, race...'' &
``...prohibit language...'' &
``...that dehumanizes others...'' &
\begin{center}-\end{center}  &
\begin{center}-\end{center}  \\

Reddit~\cite{reddit_hate_speech} &
``...groups based on their actual and perceived race, color, religion, national origin...'' &
``...attacking marginalized or vulnerable groups...'' &
``...describing a racial minority as sub-human and inferior...'' &
``...declaring that it is sickening that people of color have the right to vote...'' &
``...minimizing the scale of a hate-based violent...'' \\

TikTok~\cite{tiktok_hate_speech} &
``...attacks people based on protected attributes like race, religion, gender, or sexual orientation...'' &
``...hateful behavior...based on protected attributes...'' &
``Claiming that a protected group is physically or mentally inferior...'' &
``...demeans or furthers exclusion of protected groups...'' &
``...violent and hateful organizations or individuals...'' \\

\bottomrule
\end{tabular}
\caption{Comparison of hate speech definitions across international organizations and platform policies using explicit textual evidence.
Each cell summarizes how a given source operationalizes a core factor of hate speech.
Despite differences in wording and enforcement thresholds, all sources converge on protected-group targeting combined with hostile content, with escalating emphasis on dehumanization, exclusion, and incitement.}
\label{tab:hate_definitions_textual}
\end{table*}

\paragraph{Core definitional conditions.}
Across contexts, two conditions consistently emerge as necessary for identifying hate speech.

\begin{enumerate}
    \item[\texttt{q1})] \emph{Protected target identification.} Whether the text targets an individual or group on the basis of an inherent or protected characteristic is foundational. Without a protected target, the expression may be abusive or offensive, but it does not meet the basic definition of hate speech. Targeting may be explicit or implicit, and it may apply to individuals addressed as representatives of a group. This question therefore \emph{might act} as a gate condition for the entire checklist.

    \item[\texttt{q2})] \emph{Derogatory or hostile content.} Even when a protected group is mentioned, hate speech requires that the content convey hostility, contempt, or animus toward that group. It operationalizes this requirement by checking for derogatory tone, insults, accusations, or negative generalizations tied to the protected identity.
\end{enumerate}

\paragraph{Realizations of hateful content.}
The second group decomposes the broad notion of “attacking content” into recurrent linguistic and rhetorical realizations. Beyond the core conditions, hate speech spans a spectrum of severity. At lower levels, it includes identity-based insults, slurs, demeaning generalizations, and portrayals of inferiority. At higher levels, it encompasses dehumanization, scapegoating narratives, calls for exclusion or discrimination, threats, and explicit or implicit incitement to violence. The accumulation of these factors typically signals increasing severity.

\begin{enumerate}
    \item[\texttt{q3})] \emph{Explicit slur usage.}
    Slurs and epithets are among the strongest and most unambiguous markers of identity-based hostility.
    Their presence is often sufficient for content to be labeled as hate speech in moderation settings, regardless of additional context. We isolate this signal explicitly, while allowing other questions to capture non-slur-based hostility.
    
    \item[\texttt{q4})] \emph{Dehumanization or vilification.}
    Dehumanizing language represents a particularly dangerous form of hate speech, as it historically precedes and legitimizes violence. It captures this escalation by detecting portrayals that strip the target group of human status or depict it as inherently monstrous or threatening.
    
    \item[\texttt{q5})] \emph{Blame or scapegoating.}
    Scapegoating narratives attribute societal problems or wrongdoing to a group solely on the basis of identity. Such claims are central to many hate ideologies and serve to rationalize hostility. This question captures this mechanism, which often functions as a bridge between derogatory speech and calls for exclusion or harm.
\end{enumerate}

\paragraph{Severity and escalation signals.}
The third group captures action-oriented dimensions that increase the severity and risk associated with hate speech.

\begin{enumerate}
    \item[\texttt{q6})] \emph{Advocacy of discrimination or exclusion.}
    This question checks whether the text promotes denying rights, segregating, or excluding the group from social participation.
    Such advocacy constitutes hate speech even in the absence of explicit violence, as it encourages discriminatory treatment tied to identity.
    
    \item[\texttt{q7})] \emph{Threats or wishes of harm.}
    Expressions that articulate a desire for physical harm or death, even without a direct call to action, represent a clear escalation. It captures both explicit threats and implicit wishes of eradication.
    
    \item[\texttt{q8})] \emph{Incitement or endorsement of violence.}
    This question isolates the most severe category, where the speaker encourages or legitimizes violent action by others.
    Unlike \texttt{q7}, which captures desire or threat, \texttt{q8} focuses on mobilization and endorsement, including implicit encouragement of violent acts.
\end{enumerate}

\paragraph{Contextual and impact-based qualifiers.}
A further source of complexity lies in the context. Hateful expressions may appear in quotation, satire, reporting, or educational discussion without being endorsed by the speaker. Conversely, indirect or euphemistic language may convey hostility even in the absence of explicit slurs. Because intent is rarely observable directly, practical definitions rely on whether the content is reasonably understood as endorsing hostility or harm, and on its likely impact on the targeted group. Failing to account for context leads to systematic false positives, particularly in analytical or condemnatory uses of hateful language. The final group addresses these cases where surface cues alone are insufficient.

\begin{enumerate}
    \item[\texttt{q9})] \emph{Speaker endorsement versus quotation or satire.}
    Hateful language may appear in texts that quote, report, condemn, or analyze hate speech rather than endorse it. It evaluates whether the speaker appears to be genuinely advocating the hateful message, as opposed to using it in a critical or descriptive context.
    This question is essential for avoiding systematic false positives.
    
    \item[\texttt{q10})] \emph{Likely perceived harm to the target group.}
    Finally, we add a question that evaluates the expected impact of the content from the perspective of a reasonable member of the targeted group.
    This question captures an impact-oriented view commonly used in moderation settings, complementing intent-based assessments.
    If members of the group would reasonably feel attacked, threatened, or demeaned purely because of their identity, the content satisfies this condition.
\end{enumerate}

\subsection{Prompt construction, LLM inference and binary representation}
\label{sec:prompt_inference}

\paragraph{Prompt construction}
Each checklist question is instantiated through a prompt designed to elicit a binary judgment for one conceptual factor. Rather than relying on a generic instruction, prompts are explicitly structured to constrain both semantic interpretation and output format, while avoiding any exposure to the final hate speech label.

Our prompts have three main components.
First, the \textbf{question} explicitly states the factor being evaluated, using concise and unambiguous language.
Second, a short \textbf{rationale} accompanies the question to clarify its intended scope, disambiguate borderline cases, and align the model’s reasoning with the conceptual definition of the factor.
Third, a small set of \textbf{few-shot examples} is provided to anchor both interpretation and output behavior.
These examples include positive and negative cases, ensuring that the model learns not only what constitutes a positive instance of the factor, but also what does not.
In addition, the few-shot examples encourage the model to provide a brief natural-language justification before committing to the final answer, and reinforce the required output format for us to parse. The messages and output format we use are the following:

\noindent
\begin{minipage}[t]{0.48\textwidth}
\begin{mdframed}[
  linewidth=1.2pt,
  roundcorner=3pt,
  backgroundcolor=promptbg1,
]
\small
\begin{center}
    \textbf{\underline{System message}}
\end{center}

\vspace{1em}
You are an expert in hate speech detection.

\vspace{1pt}
Your \textbf{task} is to answer the following \textbf{question} for the given \textbf{text}:
\begin{quote}
\emph{\{question\}}
\end{quote}

\vspace{1pt}
For internal consistency, here you have a guiding \textbf{rationale} to help you answer the question:
\begin{quote}
\emph{\{rationale\}}
\end{quote}

\vspace{1pt}
Please \textbf{output} your answer at the end in the format
\texttt{<a>Yes</a>} or \texttt{<a>No</a>}.
\vspace{3.6em}
\end{mdframed}
\end{minipage}
\hfill
\begin{minipage}[t]{0.48\textwidth}

\begin{mdframed}[
  linewidth=1.2pt,
  roundcorner=3pt,
  backgroundcolor=promptbg2,
]
\small
\begin{center}
\textbf{\underline{User message}}
\end{center}

\vspace{1em}
\textbf{Text:}
\begin{quote}
\emph{\{text\}}
\end{quote}

\vspace{1pt}
Answer the \textbf{question}:

\begin{quote}
\emph{\{question\}}
\end{quote}

\vspace{1pt}
Please \textbf{output} your answer in the format \texttt{<a>Yes</a>} or \texttt{<a>No</a>}.
\end{mdframed}

\vspace{0.5em}

\begin{mdframed}[
  linewidth=1.2pt,
  roundcorner=3pt,
  backgroundcolor=promptbg3,
]
\small
\begin{center}
\textbf{\underline{Output format}}
\end{center}

\vspace{1em}

\emph{\{Short natural-language justification\}}

\vspace{2pt}

\texttt{<a>Yes|No</a>}
\end{mdframed}

\end{minipage}

Importantly, the prompt never asks the model to decide whether the input constitutes hate speech. The final label is neither mentioned nor implied. The model is therefore restricted to answering a single diagnostic question, preventing shortcut reasoning and leakage from downstream objectives.

\paragraph{Binary diagnostic representation.}
Our prompt constrains the LLM to produce a strictly binary output in the form \texttt{Yes} or \texttt{No}. Given an input text $t$, the checklist therefore induces a binary vector
\[
\mathbf{z}(t) = \big( \mathcal{I}_{q_1}(t), \mathcal{I}_{q_2}(t), \dots, \mathcal{I}_{q_{10}}(t) \big),
\]
where $\mathcal{I}_{q_i} : \mathcal{T} \rightarrow \{0,1\}$ denotes an independent LLM-based inference function answering question $q_i$.

The diagnostic space is a discrete space of size $2^{10}=1024$ possible configurations. Unlike continuous embeddings, this space has an explicit semantic interpretation: each dimension corresponds to a clearly defined conceptual factor, and each vector encodes which factors are present or absent in the input. As a result, the representation supports direct inspection, rule-based reasoning, and analysis of factor interactions.

\paragraph{Model-agnostic inference and parallelism.}
The proposed framework is model-agnostic: any instruction-following LLM capable of producing short, constrained outputs can be used. At inference time, each input text is processed by issuing one independent LLM call per checklist question. For a single input, this results in $K=10$ independent inferences, one per diagnostic factor.

No conversational state or shared context is maintained across questions.
Each prompt is executed in isolation, and the model does not have access to the answers of other questions. This design choice is deliberate. First, it prevents cross-question contamination, where a strong signal in one factor could implicitly bias responses to others. Second, it allows the parallel execution of the inferences.

\paragraph{Robust binary resolution via log-probability forcing}
In practice, not all LLM outputs can be reliably parsed into a valid binary answer.
This issue is especially pronounced for smaller or less instruction-aligned models.
We observe two main failure modes:
(i) the model refuses to answer due to internal safety or moderation mechanisms triggered by hate-related content, and
(ii) the model provides a relevant explanation but fails to conform to the required output format, making the binary decision unparseable.

Discarding such cases would introduce systematic bias and disproportionately penalize smaller models.
To address this, we introduce a lightweight solver that forces a binary resolution using token-level log-probabilities.
When an invalid or missing answer is detected, we take the raw text generated by the model and append the opening token of the expected answer format (i.e., \texttt{<a>}).
Conditioned on this augmented context, we then query the model for the log-probabilities of the candidate tokens corresponding to \texttt{Yes} and \texttt{No}.
Formally, given a prompt $p$ and model output $o$, we compute
\[
\hat{z} = \arg\max_{y \in \{\texttt{Yes}, \texttt{No}\}} \log P(y \mid \mathrm{concat} (o, \texttt{<a>})).
\]
This procedure forces the model to commit to one of the two admissible outcomes, even when it initially avoided doing so explicitly.


\subsection{Feature aggregation}
\label{sec:tree_aggregation}

Given the diagnostic representation $\mathbf{z}(t)\in\{0,1\}^{10}$ produced by the checklist, we cast the final hate speech decision as a supervised learning problem over a low-dimensional, discrete feature space.
We train a classifier $f_{\theta}$ that maps checklist outputs to dataset labels:
\[
\hat{y} = f_{\theta}(\mathbf{z}(t)), \quad \mathbf{z}(t)\in\{0,1\}^{10}, \ \hat{y}\in\{0,1\}.
\]
This aggregation layer is intentionally lightweight and fully interpretable, matching the design goal of separating semantic judgment (handled by the LLM) from decision policy (learned from data).

\paragraph{Decision tree choice.}
A decision tree is a natural choice for three reasons. First, it provides \textbf{explicit and interpretable reasoning paths}: each prediction can be explained as a conjunction of satisfied predicates (e.g., \texttt{q1}=1, \texttt{q2}=1, \texttt{q8}=1), closely mirroring the logical interpretation of the checklist. Second, \textbf{training and inference is fast and low-resource}, and negligible compared to LLM inference or fine-tuning. This does not harm performance, as the model does not require high-capacity parameterization to operate effectively on a 10-bit input. Third, it aligns with the \textbf{diagnostic objective}: the learned splits expose which conceptual factors are treated as most informative under a given dataset’s labeling scheme.

\paragraph{Hard-coded rules versus learned aggregation.}

Although the checklist admits a natural logical interpretation, we deliberately avoid enforcing a fixed aggregation rule.
The primary reason is that hate speech datasets encode heterogeneous and often implicit operational definitions, shaped by annotation guidelines, cultural context, and annotator subjectivity.
Hard-coding a single decision rule (e.g., requiring \texttt{q1} as a strict gate or treating \texttt{q7}--\texttt{q8} as sufficient conditions) would impose a normative definition that may systematically misalign with the labeling criteria of a given dataset.

Instead, we treat the checklist as a stable diagnostic representation and delegate aggregation to a lightweight classifier that can adapt to dataset-specific annotation regimes.
This separation allows the model to learn which combinations of factors are treated as decisive under different labeling standards, while preserving interpretability at the factor level.
Crucially, the learned aggregation does not obscure the decision process: the resulting decision tree still implements an explicit conjunction of checklist predicates, making the effective decision logic fully transparent and inspectable.

\section{Experimental Setup}
\label{sec:experimental_setup}

\subsection{Datasets}
We evaluate our approach on multiple publicly available hate speech datasets that differ in size, domain, annotation guidelines, and conceptual scope. All experiments are framed as a binary classification task (hate speech vs.\ non-hate speech), enabling a consistent evaluation across datasets. To assess robustness and transferability, we adopt both in-domain and cross-dataset evaluation settings.

\begin{enumerate}
\item \textbf{Measuring Hate Speech}~\cite{measuring_hate_speech} is the largest dataset used in our experiments and provides a broad and semantically rich definition of hate speech. It covers a wide range of targets, linguistic forms, and levels of severity. Due to its size and diversity, we use this dataset both for training and evaluation.

\item \textbf{HateXplain}~\cite{hatexplain} contains social media posts annotated for hate speech, abusive language, and offensive content, with additional rationales provided by annotators. We collapse the labels into a binary hate vs.\ non-hate setting following common practice. HateXplain is used both for training and evaluation and provides a complementary annotation style that emphasizes explanation and human reasoning.

\item The \textbf{Stormfront} dataset~\cite{stormfront} consists of forum posts from a white supremacist online community and contains explicit and implicit forms of hate speech. Its language is often more extreme and domain-specific. We include Stormfront both as a training and evaluation dataset to test performance on highly polarized and ideologically charged content.

\item \textbf{ETHOS}~\cite{ethos} is a smaller dataset focusing on explicit and implicit hate speech, particularly in short social media texts. Due to its limited size, we use ETHOS exclusively for evaluation, allowing us to assess cross-dataset generalization.

\item \textbf{HateCheck}~\cite{hatecheck} is a functional test suite rather than a naturally occurring dataset. It consists of carefully constructed examples designed to probe specific capabilities and failure modes of hate speech detection systems, such as handling negation, counter-speech, or protected group substitutions. We use HateCheck only for evaluation.
\end{enumerate}

Across all datasets, we follow the original train--test splits when available. In cross-dataset experiments, models are trained on one dataset and evaluated on others without any additional adaptation.

\subsection{Baselines}
\label{subsec:baselines}

We compare our proposed diagnostic framework against two strong and commonly adopted baselines: zero-shot LLM classification and in-domain supervised fine-tuning.
To ensure comparability across datasets and models, we evaluate all methods using the same metric and experimental protocol.

\begin{description}
    \item[Zero-shot LLM classification.]In this baseline, decoder-based language models are prompted to predict whether a given input text constitutes hate speech. The prompt explicitly instructs the model to output a binary label (\texttt{Yes} or \texttt{No}), without access to the diagnostic checklist or any task-specific training examples. This setting reflects a common real-world usage of LLMs for content moderation, where models are deployed without dataset-specific fine-tuning or calibration.

    \item[In-domain supervised fine-tuning.]
    As a second baseline, we fine-tune all models using standard supervised learning on each dataset.
    All models are initialized from their \emph{base} (non-instruction-tuned) checkpoints, ensuring that performance gains stem from task-specific supervision rather than prior alignment.
    
    For encoder-based models, we attach a linear classification head to the pooled representation of the \texttt{[CLS]} token.
    For decoder-based LLMs, we train a lightweight classification head on top of the hidden state of the final token in the input sequence, and predict the hate vs.\ non-hate label from that representation.
    This design keeps the fine-tuning setup consistent across architectures while respecting their representational differences.
    
    Fine-tuning is performed in-domain, using the standard train--test split of each dataset.
    Across all models and datasets, we observe stable convergence within five epochs, with early stopping applied based on validation performance.
    All models are trained using a shared set of hyperparameters to ensure a controlled comparison.
    \Cref{tab:training_hyperparameters} summarizes the most relevant settings.
    Unless otherwise noted, these parameters are kept fixed across architectures and datasets.
    
    \begin{table}[!ht]
\centering
\footnotesize
\setlength{\tabcolsep}{6pt}
\renewcommand{\arraystretch}{1.15}

\begin{tabular}{rl|rl}
\toprule
\textbf{Hyperparameter} & \textbf{Value} & \textbf{Hyperparameter} & \textbf{Value} \\
\midrule
Max Sequence Length & 512 & Train Batch Size & 32 \\
Optimizer & AdamW (fused) & LR Scheduler & Linear \\
Learning Rate & $2 \times 10^{-5}$ & Warmup Ratio & 0.1 \\
Weight Decay & 0.01 & LoRA Rank ($r$) & 8 \\
Number of Epochs & 5 & LoRA Scaling ($\alpha$) & 32 \\
Early Stopping Patience & 2 & LoRA Dropout & 0.1 \\
\bottomrule
\end{tabular}


\vspace{5pt} 
\caption{Shared fine-tuning hyperparameters used across all models.
All experiments start from base (non-instruction-tuned) checkpoints.
With these settings, all models reliably converge within five epochs.}
\label{tab:training_hyperparameters}
\end{table}
    
    To enable efficient fine-tuning of large decoder-based models, we employ parameter-efficient adaptation.
    Specifically, we use Low-Rank Adaptation (LoRA)~\cite{lora} for all decoder models, injecting trainable rank-$r$ updates into the attention layers while keeping the base model weights frozen.
    
    For models with more than $8$B parameters, we further adopt QLoRA training, loading the base model in $4$-bit quantized form and applying LoRA adapters on top~\cite{qlora}.
    This substantially reduces memory requirements while preserving fine-tuning quality, allowing us to train models up to $27$B parameters under a unified setup.
    All encoder-only and small LLMs models are fine-tuned without parameter-efficient adapters, following standard full-parameter training.
    
    
\end{description}

\subsection{Model Selection}
\label{subsec:model_selection}
To ensure a fair and comprehensive evaluation, we experiment with a diverse set of language models that span different architectures and parameter scales. This allows us to study how model capacity and design choices affect zero-shot performance, supervised learning, and our diagnostic approach.

\begin{itemize}
    \item \textbf{BERT-type encoder models.}
    We include several widely used encoder-only transformer models that represent the standard backbone for supervised hate speech classification. Specifically, we evaluate BERT-base~\cite{bert}, RoBERTa-base~\cite{roberta}, DeBERTa-base~\cite{debertav3}, and ModernBERT-base~\cite{modernbert}. These models are only used in supervised fine-tuning settings, as they do not support zero-shot classification.

    \item \textbf{Decoder LLMs.} We evaluate decoder-based models across three scales: \textbf{Small} (LLaMA~3~1B/3B, Gemma~3~4B) for resource-constrained inference; \textbf{Mid-size} (Mistral~7B, LLaMA~3~8B, Gemma~3~12B) for balanced performance; and \textbf{Large} (Mistral~24B, Gemma~3~27B). The particular models are shown in \Cref{tab:model_specs}, together with the quantization method used and the estimated size.
    
    \begin{table*}[!ht]
\centering
\small
\setlength{\tabcolsep}{10pt}
\renewcommand{\arraystretch}{1.2}
\begin{tabular}{lccc}
\toprule
\textbf{Model (Hugging Face ID)} & \textbf{Quantization} & \textbf{Parameters} & \textbf{Est. Size} \\
\midrule
\textit{meta-llama/Llama-3.2-1B-Instruct} & None (BF16) & 1B & $\sim$2.47 GB \\
\textit{meta-llama/Llama-3.2-3B-Instruct} & None (BF16) & 3B & $\sim$6.43 GB \\
\textit{google/gemma-3-4b-it} & None (BF16) & 4B & $\sim$8.60 GB \\
\textit{mistralai/Mistral-7B-Instruct-v0.3} & None (BF16) & 7B & $\sim$14.50 GB \\
lurker18/Llama\_3.1\_8B\_Instruct\_AWQ\_4bit & 4-bit (AWQ) & 8B & $\sim$5.73 GB \\
\textit{gaunernst/gemma-3-12b-it-int4-awq} & 4-bit (AWQ) & 12B & $\sim$8.97 GB \\
\textit{stelterlab/Mistral-Small-24B-Instruct-2501-AWQ} & 4-bit (AWQ) & 24B & $\sim$14.24 GB \\
\textit{gaunernst/gemma-3-27b-it-int4-awq} & 4-bit (AWQ) & 27B & $\sim$18.46 GB \\
\bottomrule
\end{tabular}
\caption{Model specifications including quantization format and estimated disk/VRAM footprint.}
\label{tab:model_specs}
\end{table*}
    

    

\end{itemize}

Across all settings, the same set of models is used consistently to isolate the effect of training strategy and inference methodology rather than model choice.

\subsection{Checklist feature aggregator: decision tree configuration}
For the decision tree operating on top of the checklist outputs, we deliberately adopt an extremely simple experimental setup.
Our goal is to reflect the explanatory power of the diagnostic representation itself, rather than gains obtained through extensive model tuning.
Accordingly, we do not perform any hyperparameter optimization for the decision tree.

We use the standard \texttt{DecisionTreeClassifier} implementation from \texttt{scikit-learn}~\footnote{\url{https://scikit-learn.org/stable/modules/generated/sklearn.tree.DecisionTreeClassifier.html}}, with all parameters left at their default values except for \texttt{min\_weight\_fraction\_leaf}, which is set to $0.01$.
This parameter enforces that each leaf node must contain at least $1\%$ of the total sample weight, preventing the tree from creating leaves that explain only a handful of training examples.


\subsection{Evaluation metric}
We report performance primarily in terms of the Area Under the ROC Curve (AUC).
This choice is motivated by two considerations.
First, we mostly test out-of-distribution performance, we focus on the ability of the models to separate positive and negative class regardless of the particular threshold of choice, which can be re-trained on a small amount of data.
Secondly, hate speech datasets often exhibit substantial class imbalance, for which threshold-dependent metrics such as accuracy can be misleading.

For zero-shot LLM classification, we derive a scalar prediction score from the normalized log-probabilities assigned to the answer tokens \texttt{Yes} and \texttt{No}.
Specifically, given the model’s output distribution conditioned on the prompt and justification, we compute the normalized probability of the positive class and use this value to construct the ROC curve.

In addition to absolute AUC, we report a \emph{relative AUC} metric to explicitly quantify cross-dataset robustness.
Formally, let $\mathcal{D}_1$ and $\mathcal{D}_2$ be two different datasets, and let $M(\mathcal{D}^{\text{train}})$ denote a model fitted to the training split of the dataset $\mathcal{D}$. Then
\[
\mathrm{RelAUC}(M; \mathcal{D}_{1}, \mathcal{D}_{2}) \;=\;
\frac{\mathrm{AUC}(M(\mathcal{D}_1^{\text{train}}); \mathcal{D}_2^{\text{test}})}{\mathrm{AUC}(M(\mathcal{D}_1^{\text{train}}); \mathcal{D}_1^{\text{test}})} \times 100.
\]

This normalization controls for differences in intrinsic dataset difficulty and highlights how well performance transfers relative to in-domain performance.
A relative AUC close to or above $100\%$ indicates strong robustness under domain shift, whereas lower values reflect degradation when moving away from the training distribution.

Finally, we report an \emph{out-of-domain AUC} (OOD AUC) to summarize cross-dataset generalization in a single scalar.
Given a model $M$ trained on a source dataset $\mathcal{D}^{\text{train}}$ and evaluated on a set of $K$ distinct target datasets
$\{\mathcal{D}_1, \dots, \mathcal{D}_K\}$ distinct from the source one, we define the OOD AUC as
\[
\mathrm{AUC}_{\text{OOD}}(M; D) \;=\;
\frac{1}{K} \sum_{k=1}^{K} \mathrm{RelAUC}(M; \mathcal{D}, \mathcal{D}_{k})
.\]
This metric provides a compact estimate of a model’s average performance under domain shift, abstracting away from any single target dataset.


\subsection{Implementation details and benchmarking}
\label{subsec:efficient_inference}

\begin{table*}[!ht]
\centering
\small
\setlength{\tabcolsep}{8pt}
\renewcommand{\arraystretch}{1.2}

\begin{tabular}{
l
c
c
c
}
\toprule
\textbf{Model (Hugging Face ID)} &
\textbf{\makecell{Checklist inference \\ time per sample (s)}} &
\textbf{\makecell{Checklist inference \\ time on MHS}} &
\textbf{\makecell{Supervised \\ training time on MHS}} \\
\midrule

\textit{meta-llama/Llama-3.2-1B-Instruct}%
  & 0.167 & 1h 50min & 50min \\

\textit{meta-llama/Llama-3.2-3B-Instruct}%
  & 0.309 & 3h 20min & 2h 10min \\

\textit{google/gemma-3-4b-it}%
  & 0.187 & 2h & 2h 30min \\

\textit{mistralai/Mistral-7B-Instruct-v0.3}%
  & 0.625 & 6h 50min & 5h 50min \\

\textit{lurker18/Llama\_3.1\_8B\_Instruct\_AWQ\_4bit}%
  & 0.458 & 5h & 2h 50min \\

\textit{gaunernst/gemma-3-12b-it-int4-awq}%
  & 0.709 & 7h 40min & 15h 10min \\

\textit{stelterlab/Mistral-Small-24B-Instruct-2501-AWQ}%
  & 1.238 & 13h 30min & 29h 40min \\

\textit{gaunernst/gemma-3-27b-it-int4-awq}%
  & 2.704 & 29h 40min & 35h 20min \\

\bottomrule
\end{tabular}

\caption{Computational cost comparison between checklist-based inference and supervised fine-tuning on the Measuring Hate Speech dataset (39568 samples).
Checklist inference time per sample reports the end-to-end latency required to answer all ten checklist questions for a single input.}
\label{tab:generation_times}
\end{table*}

All inference experiments are conducted on a single \textit{NVIDIA RTX 4500 Ada Generation} GPU with 24\,GB of VRAM. All generations, both for the zero-shot baseline and for the checklist-based method, rely on the \texttt{vLLM} inference library.
\texttt{vLLM} provides highly optimized decoding through continuous batching, efficient KV-cache management, and low-overhead scheduling~\footnote{\url{https://vllm.ai/}}.

\paragraph{Inference efficiency across models.}
\Cref{tab:generation_times} reports the computational cost of checklist-based inference across all evaluated LLMs.
Although the checklist requires ten independent generations per input, the overall runtime is comparable to the cost of supervised fine-tuning on the same dataset, particularly for mid- and large-scale models.

However, for the checklist approach, once the diagnostic responses are generated, training the downstream decision tree incurs negligible overhead.
In contrast, supervised methods require full retraining when adapting to new data or annotation regimes, which becomes increasingly expensive as model size grows.

This makes the checklist framework particularly scalable.
Incorporating additional data simply requires generating checklist responses for the new samples and retraining a lightweight classifier, while the supervised setting typically demands re-training the full model or applying non-trivial continual learning strategies.
As a result, the proposed method offers a favorable trade-off between computational cost, robustness, and interpretability, especially in settings where datasets evolve or multiple annotation schemes must be supported.

\section{Results}
\label{sec:results}



\begin{table*}[!t]
\centering
\small
\setlength{\tabcolsep}{2.75pt}
\renewcommand{\arraystretch}{1.25}

\begin{tabularx}{\textwidth}{
l l |
Y Y Y Y Y Y Y Y Y Y
}
\toprule
\multirow{2}{*}{\textbf{Model}} &
\multirow{2}{*}{\textbf{Approach}} &
\multicolumn{2}{c}{\textbf{ETHOS}} &
\multicolumn{2}{c}{\textbf{HateCheck}} &
\multicolumn{2}{c}{\textbf{HateXplain}} &
\multicolumn{2}{c}{\textbf{MHS}} &
\multicolumn{2}{c}{\textbf{Stormfront}} \\
\cmidrule(lr){3-4}\cmidrule(lr){5-6}\cmidrule(lr){7-8}\cmidrule(lr){9-10}\cmidrule(lr){11-12}
& &
\textbf{AUC} & \textbf{$\Delta$} &
\textbf{AUC} & \textbf{$\Delta$} &
\textbf{AUC} & \textbf{$\Delta$} &
\textbf{AUC} & \textbf{$\Delta$} &
\textbf{AUC} & \textbf{$\Delta$} \\
\midrule

\rowcolor{llama!50}
  & zero-shot
  & 0.713 & 
  & 0.718 & 
  & 0.534 & 
  & 0.599 & 
  & 0.726 & \\
\rowcolor{llama!50}
\multirow{-2}{*}{\texttt{\mbox{llama3-1b}}}
  & checklist
  & 0.724 & \multirow{-2}{*}{\deltacell{+0.011}}
  & 0.743 & \multirow{-2}{*}{\deltacell{+0.024}}
  & 0.519 & \multirow{-2}{*}{\deltacell{-0.015}}
  & 0.634 & \multirow{-2}{*}{\deltacell{+0.035}}
  & 0.708 & \multirow{-2}{*}{\deltacell{-0.018}} \\

\rowcolor{llama!100}
  & zero-shot
  & 0.804 & 
  & 0.783 & 
  & 0.621 & 
  & 0.691 & 
  & 0.765 & \\
\rowcolor{llama!100}
\multirow{-2}{*}{\texttt{\mbox{llama3-3b}}}
  & checklist
  & 0.883 & \multirow{-2}{*}{\deltacell{+0.079}}
  & 0.898 & \multirow{-2}{*}{\deltacell{+0.115}}
  & 0.767 & \multirow{-2}{*}{\deltacell{+0.146}}
  & 0.815 & \multirow{-2}{*}{\deltacell{+0.124}}
  & 0.850 & \multirow{-2}{*}{\deltacell{+0.084}} \\

\rowcolor{gemma!50}
  & zero-shot
  & 0.819 & 
  & 0.812 & 
  & 0.619 & 
  & 0.671 & 
  & 0.797 & \\
\rowcolor{gemma!50}
\multirow{-2}{*}{\texttt{\mbox{gemma3-4b}}}
  & checklist
  & 0.882 & \multirow{-2}{*}{\deltacell{+0.063}}
  & 0.880 & \multirow{-2}{*}{\deltacell{+0.069}}
  & 0.806 & \multirow{-2}{*}{\deltacell{+0.187}}
  & 0.819 & \multirow{-2}{*}{\deltacell{+0.148}}
  & 0.860 & \multirow{-2}{*}{\deltacell{+0.064}} \\

\rowcolor{mistral!100}
  & zero-shot
  & 0.853 & 
  & 0.835 & 
  & 0.621 & 
  & 0.696 & 
  & 0.800 & \\
\rowcolor{mistral!100}
\multirow{-2}{*}{\texttt{\mbox{mistral-7b}}}
  & checklist
  & 0.908 & \multirow{-2}{*}{\deltacell{+0.054}}
  & 0.935 & \multirow{-2}{*}{\deltacell{+0.100}}
  & 0.797 & \multirow{-2}{*}{\deltacell{+0.176}}
  & 0.811 & \multirow{-2}{*}{\deltacell{+0.115}}
  & 0.877 & \multirow{-2}{*}{\deltacell{+0.076}} \\

\rowcolor{llama!150}
  & zero-shot
  & 0.799 & 
  & 0.803 & 
  & 0.562 & 
  & 0.648 & 
  & 0.795 & \\
\rowcolor{llama!150}
\multirow{-2}{*}{\texttt{\mbox{llama3-8b}}}
  & checklist
  & 0.804 & \multirow{-2}{*}{\deltacell{+0.005}}
  & 0.853 & \multirow{-2}{*}{\deltacell{+0.050}}
  & 0.608 & \multirow{-2}{*}{\deltacell{+0.046}}
  & 0.725 & \multirow{-2}{*}{\deltacell{+0.077}}
  & 0.784 & \multirow{-2}{*}{\deltacell{-0.011}} \\

\rowcolor{gemma!100}
  & zero-shot
  & 0.845 & 
  & 0.843 & 
  & 0.662 & 
  & 0.696 & 
  & 0.813 & \\
\rowcolor{gemma!100}
\multirow{-2}{*}{\texttt{\mbox{gemma3-12b}}}
  & checklist
  & 0.902 & \multirow{-2}{*}{\deltacell{+0.057}}
  & 0.929 & \multirow{-2}{*}{\deltacell{+0.086}}
  & 0.831 & \multirow{-2}{*}{\deltacell{+0.170}}
  & 0.762 & \multirow{-2}{*}{\deltacell{+0.066}}
  & 0.866 & \multirow{-2}{*}{\deltacell{+0.054}} \\

\rowcolor{mistral!150}
  & zero-shot
  & 0.834 & 
  & 0.877 & 
  & 0.625 & 
  & 0.686 & 
  & 0.815 & \\
\rowcolor{mistral!150}
\multirow{-2}{*}{\texttt{\mbox{mistral-24b}}}
  & checklist
  & 0.912 & \multirow{-2}{*}{\deltacell{+0.078}}
  & 0.956 & \multirow{-2}{*}{\deltacell{+0.079}}
  & 0.802 & \multirow{-2}{*}{\deltacell{+0.178}}
  & 0.796 & \multirow{-2}{*}{\deltacell{+0.109}}
  & 0.921 & \multirow{-2}{*}{\deltacell{+0.106}} \\

\rowcolor{gemma!150}
  & zero-shot
  & 0.832 & 
  & 0.868 & 
  & 0.605 & 
  & 0.679 & 
  & 0.806 & \\
\rowcolor{gemma!150}
\multirow{-2}{*}{\texttt{\mbox{gemma3-27b}}}
  & checklist
  & 0.902 & \multirow{-2}{*}{\deltacell{+0.070}}
  & 0.925 & \multirow{-2}{*}{\deltacell{+0.057}}
  & 0.802 & \multirow{-2}{*}{\deltacell{+0.197}}
  & 0.779 & \multirow{-2}{*}{\deltacell{+0.107}}
  & 0.877 & \multirow{-2}{*}{\deltacell{+0.071}} \\

\bottomrule
\end{tabularx}

\caption{Comparison between direct zero-shot LLM classification and the proposed \texttt{checklist} approach across multiple datasets.
For the \texttt{checklist} approach, the reported AUC on dataset \texttt{X} is computed by averaging the performance across training in all datasets \emph{except} \texttt{X}.
This design choice mitigates in-domain bias and enables a fairer comparison between zero-shot prediction and the checklist-based framework.
$\Delta$ denotes the absolute AUC difference between the checklist and zero-shot approaches for each dataset.
}
\label{tab:checklist_vs_zero-shot}
\end{table*}

\begin{table*}[!ht]
\centering
\small
\setlength{\tabcolsep}{2.75pt}
\renewcommand{\arraystretch}{1.25}

\begin{tabularx}{\textwidth}{
l l |
c c c c c c c c c |
c c c
}
\toprule

\multirow{2}{*}{\textbf{Model}} &
\multirow{2}{*}{\textbf{Approach}} &
\multicolumn{1}{c}{\textbf{MHS}} &
\multicolumn{2}{c}{\textbf{HateXplain}} &
\multicolumn{2}{c}{\textbf{Stormfront}} &
\multicolumn{2}{c}{\textbf{ETHOS}} &
\multicolumn{2}{c|}{\textbf{HateCheck}} &
\multirow{2}{*}{\textbf{\makecell{Rel.\\AUC}}} &
\multirow{2}{*}{\textbf{\makecell{Avg.\\AUC}}} &
\multirow{2}{*}{\textbf{\makecell{OOD\\AUC}}} \\
\cmidrule(lr){3-3}\cmidrule(lr){4-5}\cmidrule(lr){6-7}\cmidrule(lr){8-9}\cmidrule(lr){10-11}
& &
\textbf{AUC} &
\textbf{AUC} & \textbf{(\%)} &
\textbf{AUC} & \textbf{(\%)} &
\textbf{AUC} & \textbf{(\%)} &
\textbf{AUC} & \textbf{(\%)} &
& & \\
\midrule

\rowcolor{llama!50}
 & checklist  & 0.642 & 0.600 & 93.58\% & 0.714 & 111.35\% & 0.698 & 108.74\% & 0.722 & 120.30\% & \underline{\textbf{108.49\%}} & 0.675 & 0.684 \\
\rowcolor{llama!50}
\multirow{-2}{*}{\texttt{\mbox{llama3-1b}}}
 & supervised & 0.894 & 0.787 & 88.02\% & 0.870 & 97.33\%  & 0.865 & 96.77\%  & 0.883 & 112.15\% & 98.57\%  & 0.860 & \underline{\textbf{0.851}} \\

\rowcolor{llama!100}
 & checklist  & 0.825 & 0.770 & 93.35\% & 0.857 & 103.85\% & 0.886 & 107.41\% & 0.898 & 116.61\% & \underline{\textbf{105.31\%}} & 0.847 & 0.853 \\
\rowcolor{llama!100}
\multirow{-2}{*}{\texttt{\mbox{llama3-3b}}}
 & supervised & 0.909 & 0.834 & 91.71\% & 0.823 & 90.54\%  & 0.875 & 96.23\%  & 0.924 & 110.83\% & 97.33\%  & 0.873 & \underline{\textbf{0.864}} \\

\rowcolor{gemma!50}
 & checklist  & 0.830 & 0.803 & 96.75\% & 0.855 & 102.99\% & 0.881 & 106.08\% & 0.876 & 109.03\% & \underline{\textbf{103.71\%}} & 0.849 & 0.854 \\
\rowcolor{gemma!50}
\multirow{-2}{*}{\texttt{\mbox{gemma3-4b}}}
 & supervised & 0.907 & 0.826 & 91.13\% & 0.892 & 98.37\%  & 0.885 & 97.59\%  & 0.941 & 113.85\% & 100.23\% & 0.890 & \underline{\textbf{0.886}} \\

\rowcolor{mistral!100}
 & checklist  & 0.838 & 0.816 & 97.39\% & 0.875 & 104.40\% & 0.908 & 108.31\% & 0.942 & 115.35\% & \underline{\textbf{106.36\%}} & 0.876 & 0.885 \\
\rowcolor{mistral!100}
\multirow{-2}{*}{\texttt{\mbox{mistral-7b}}}
 & supervised & 0.908 & 0.832 & 91.63\% & 0.882 & 97.21\%  & 0.895 & 98.62\%  & 0.951 & 114.36\% & 100.45\% & 0.894 & \underline{\textbf{0.890}} \\

\rowcolor{llama!150}
 & checklist  & 0.780 & 0.660 & 84.59\% & 0.846 & 108.48\% & 0.850 & 108.98\% & 0.908 & 137.68\% & \underline{\textbf{109.93\%}} & 0.809 & 0.816 \\
\rowcolor{llama!150}
\multirow{-2}{*}{\texttt{\mbox{llama3-8b}}}
 & supervised & 0.911 & 0.840 & 92.26\% & 0.882 & 96.85\%  & 0.880 & 96.64\%  & 0.943 & 112.26\% & 99.50\%  & 0.891 & \underline{\textbf{0.887}} \\

\rowcolor{gemma!100}
 & checklist  & 0.842 & 0.813 & 96.61\% & 0.872 & 103.64\% & 0.893 & 106.10\% & 0.904 & 111.22\% & \underline{\textbf{104.39\%}} & 0.865 & 0.871 \\
\rowcolor{gemma!100}
\multirow{-2}{*}{\texttt{\mbox{gemma3-12b}}}
 & supervised & 0.914 & 0.836 & 91.40\% & 0.888 & 97.17\%  & 0.895 & 97.85\%  & 0.964 & 115.38\% & 100.45\% & 0.899 & \underline{\textbf{0.896}} \\

\rowcolor{mistral!150}
 & checklist  & 0.842 & 0.811 & 96.26\% & 0.918 & 109.03\% & 0.912 & 108.28\% & 0.945 & 116.61\% & \underline{\textbf{107.54\%}} & 0.886 & 0.897 \\
\rowcolor{mistral!150}
\multirow{-2}{*}{\texttt{\mbox{mistral-24b}}}
 & supervised & 0.912 & 0.844 & 92.57\% & 0.917 & 100.57\% & 0.912 & 100.03\% & 0.965 & 114.27\% & 101.86\% & 0.910 & \underline{\textbf{0.910}} \\

\rowcolor{gemma!150}
 & checklist  & 0.846 & 0.806 & 95.27\% & 0.873 & 103.18\% & 0.895 & 105.77\% & 0.906 & 112.48\% & \underline{\textbf{104.18\%}} & 0.865 & 0.870 \\
\rowcolor{gemma!150}
\multirow{-2}{*}{\texttt{\mbox{gemma3-27b}}}
 & supervised & 0.913 & 0.844 & 92.41\% & 0.914 & 100.10\% & 0.891 & 97.58\% & 0.906 & 107.36\% & 99.36\% & 0.894 & \underline{\textbf{0.889}} \\

\midrule
\multicolumn{14}{l}{\textbf{Encoder-only}}\\
\addlinespace[2pt]

\rowcolor{encoder!10}
\texttt{bert}       & supervised & 0.869 & 0.707 & 81.28\% & 0.797 & 91.73\% & 0.800 & 92.08\% & 0.663 & 93.79\% & 89.72\% & 0.767 & 0.742 \\

\rowcolor{encoder!150}
\texttt{roberta}    & supervised & 0.872 & 0.725 & 83.12\% & 0.771 & 88.42\% & 0.791 & 90.64\% & 0.664 & 91.61\% & 88.45\% & 0.765 & 0.738 \\

\rowcolor{encoder!10}
\texttt{deberta}    & supervised & 0.763 & 0.678 & 88.82\% & 0.850 & 111.31\% & 0.835 & 109.35\% & 0.748 & 110.25\% & 104.93\% & 0.775 & 0.778 \\

\rowcolor{encoder!150}
\texttt{modernbert} & supervised & 0.877 & 0.703 & 80.22\% & 0.794 & 90.58\% & 0.792 & 90.29\% & 0.714 & 101.47\% & 90.64\% & 0.776 & 0.751 \\
\bottomrule
\end{tabularx}

\caption{AUC for different models comparing supervised vs. \texttt{checklist} approaches, evaluated on multiple test datasets. All models were trained on the \textbf{Measuring Hate Speech} dataset and evaluated both in-domain and out-of-domain . Percent values (\%) report the relative AUC. \textit{Rel. AUC} denotes the average relative AUC across all test datasets, \textit{Avg. AUC} the mean of the absolute AUC scores, and \textit{OOD AUC} the mean AUC computed by excluding the in-domain evaluation on MHS.}
\label{tab:auc_mhs}
\end{table*}

\begin{table*}[!ht]
\centering
\small
\setlength{\tabcolsep}{2.4pt}
\renewcommand{\arraystretch}{1.25}

\begin{tabularx}{\textwidth}{
l l |
c c c c c c c c c |
c c c
}
\toprule

\multirow{2}{*}{\textbf{Model}} &
\multirow{2}{*}{\textbf{Approach}} &
\multicolumn{1}{c}{\textbf{HateX}} &
\multicolumn{2}{c}{\textbf{MHS}} &
\multicolumn{2}{c}{\textbf{Stormfront}} &
\multicolumn{2}{c}{\textbf{ETHOS}} &
\multicolumn{2}{c|}{\textbf{HateCheck}} &
\multirow{2}{*}{\textbf{\makecell{Rel.\\AUC}}} &
\multirow{2}{*}{\textbf{\makecell{Avg.\\AUC}}} &
\multirow{2}{*}{\textbf{\makecell{OOD\\AUC}}} \\
\cmidrule(lr){3-3}\cmidrule(lr){4-5}\cmidrule(lr){6-7}\cmidrule(lr){8-9}\cmidrule(lr){10-11}
& &
\textbf{AUC} &
\textbf{AUC} & \textbf{(\%)} &
\textbf{AUC} & \textbf{(\%)} &
\textbf{AUC} & \textbf{(\%)} &
\textbf{AUC} & \textbf{(\%)} &
& & \\
\midrule

\rowcolor{llama!50}
 & checklist  & 0.626 & 0.614 & 98.19\% & 0.532 & 84.98\% & 0.557 & 88.95\% & 0.575 & 91.85\% & \underline{\textbf{91.00\%}} & 0.581 & 0.569 \\
\rowcolor{llama!50}
\multirow{-2}{*}{\texttt{\mbox{llama3-1b}}}
 & supervised & 0.911 & 0.778 & 85.39\% & 0.808 & 88.72\% & 0.798 & 87.57\% & 0.770 & 84.53\% & 86.55\% & 0.813 & \underline{\textbf{0.788}} \\

\rowcolor{llama!100}
 & checklist  & 0.775 & 0.814 & 105.13\% & 0.847 & 109.33\% & 0.878 & 113.34\% & 0.894 & 115.41\% & \underline{\textbf{110.80\%}} & 0.842 & \underline{\textbf{0.858}} \\
\rowcolor{llama!100}
\multirow{-2}{*}{\texttt{\mbox{llama3-3b}}}
 & supervised & 0.920 & 0.771 & 83.88\% & 0.815 & 88.59\% & 0.830 & 90.25\% & 0.905 & 98.45\% & 90.29\% & 0.848 & 0.830 \\

\rowcolor{gemma!50}
 & checklist  & 0.818 & 0.815 & 99.58\% & 0.866 & 105.78\% & 0.873 & 106.66\% & 0.864 & 105.51\% & \underline{\textbf{104.38\%}} & 0.847 & 0.854 \\
\rowcolor{gemma!50}
\multirow{-2}{*}{\texttt{\mbox{gemma3-4b}}}
 & supervised & 0.921 & 0.833 & 90.41\% & 0.887 & 96.29\% & 0.883 & 95.82\% & 0.934 & 101.40\% & 95.98\% & 0.892 & \underline{\textbf{0.884}} \\

\rowcolor{mistral!100}
 & checklist  & 0.826 & 0.821 & 99.37\% & 0.878 & 106.28\% & 0.908 & 109.94\% & 0.933 & 112.99\% & \underline{\textbf{107.15\%}} & 0.873 & \underline{\textbf{0.885}} \\
\rowcolor{mistral!100}
\multirow{-2}{*}{\texttt{\mbox{mistral-7b}}}
 & supervised & 0.925 & 0.794 & 85.86\% & 0.858 & 92.83\% & 0.882 & 95.36\% & 0.915 & 98.99\% & 93.26\% & 0.875 & 0.862 \\

\rowcolor{llama!150}
 & checklist  & 0.710 & 0.743 & 104.66\% & 0.723 & 101.86\% & 0.717 & 101.10\% & 0.742 & 104.62\% & \underline{\textbf{103.06\%}} & 0.727 & 0.731 \\
\rowcolor{llama!150}
\multirow{-2}{*}{\texttt{\mbox{llama3-8b}}}
 & supervised & 0.926 & 0.802 & 86.60\% & 0.747 & 80.66\% & 0.852 & 92.03\% & 0.928 & 100.15\% & 89.86\% & 0.851 & \underline{\textbf{0.832}} \\

\rowcolor{gemma!100}
 & checklist  & 0.848 & 0.786 & 92.70\% & 0.860 & 101.43\% & 0.903 & 106.41\% & 0.935 & 110.25\% & \underline{\textbf{102.70\%}} & 0.867 & \underline{\textbf{0.871}} \\
\rowcolor{gemma!100}
\multirow{-2}{*}{\texttt{\mbox{gemma3-12b}}}
 & supervised & 0.928 & 0.792 & 85.43\% & 0.848 & 91.38\% & 0.864 & 93.09\% & 0.940 & 101.36\% & 92.81\% & 0.874 & 0.861 \\

\rowcolor{mistral!150}
 & checklist  & 0.830 & 0.821 & 98.93\% & 0.923 & 111.22\% & 0.914 & 110.04\% & 0.947 & 114.01\% & \underline{\textbf{108.55\%}} & 0.887 & \underline{\textbf{0.901}} \\
\rowcolor{mistral!150}
\multirow{-2}{*}{\texttt{\mbox{mistral-24b}}}
 & supervised & 0.924 & 0.687 & 74.37\% & 0.670 & 72.53\% & 0.752 & 81.37\% & 0.524 & 56.70\% & 71.24\% & 0.711 & 0.658 \\

\rowcolor{gemma!150}
 & checklist  & 0.837 & 0.824 & 98.45\% & 0.881 & 105.18\% & 0.905 & 108.16\% & 0.922 & 110.08\% & \underline{\textbf{105.47\%}} & 0.874 & \underline{\textbf{0.883}} \\
\rowcolor{gemma!150}
\multirow{-2}{*}{\texttt{\mbox{gemma3-27b}}}
 & supervised & 0.915 & 0.823 & 89.96\% & 0.831 & 90.75\% & 0.852 & 93.13\% & 0.950 & 103.80\% & 94.41\% & 0.874 & 0.864 \\

\midrule
\multicolumn{14}{l}{\textbf{Encoder-only}}\\
\addlinespace[2pt]

\rowcolor{encoder!10}
\texttt{bert}       & supervised & 0.866 & 0.666 & 76.86\% & 0.777 & 89.64\% & 0.711 & 82.03\% & 0.600 & 69.25\% & 79.44\% & 0.724 & 0.688 \\

\rowcolor{encoder!150}
\texttt{roberta}    & supervised & 0.873 & 0.782 & 89.56\% & 0.766 & 87.69\% & 0.721 & 82.51\% & 0.568 & 65.07\% & 81.21\% & 0.742 & 0.709 \\

\rowcolor{encoder!10}
\texttt{deberta}    & supervised & 0.149 & 0.215 & 144.92\% & 0.258 & 173.50\% & 0.319 & 214.89\% & 0.358 & 241.11\% & 193.60\% & 0.260 & 0.288 \\

\rowcolor{encoder!150}
\texttt{modernbert} & supervised & 0.859 & 0.763 & 88.89\% & 0.670 & 78.00\% & 0.692 & 80.60\% & 0.593 & 69.05\% & 79.13\% & 0.715 & 0.680 \\
\bottomrule
\end{tabularx}

\caption{AUC for different models comparing supervised vs.\ \texttt{checklist} approaches, evaluated on multiple test datasets. All models were trained on HateXplain and evaluated both in-domain and out-of-domain. Percent values (\%) report the relative AUC. \textit{Rel.\ AUC} denotes the average relative AUC across all test datasets, \textit{Avg.\ AUC} the mean of the absolute AUC scores, and \textit{OOD AUC} the mean AUC computed by excluding the in-domain evaluation on HateXplain.}
\label{tab:auc_hatexplain}
\end{table*}

\begin{table*}[!ht]
\centering
\small
\setlength{\tabcolsep}{2.75pt}
\renewcommand{\arraystretch}{1.25}

\begin{tabularx}{\textwidth}{
l l |
c c c c c c c c c |
c c c
}
\toprule

\multirow{2}{*}{\textbf{Model}} &
\multirow{2}{*}{\textbf{Approach}} &
\multicolumn{1}{c}{\textbf{Storm}} &
\multicolumn{2}{c}{\textbf{MHS}} &
\multicolumn{2}{c}{\textbf{HateXplain}} &
\multicolumn{2}{c}{\textbf{ETHOS}} &
\multicolumn{2}{c|}{\textbf{HateCheck}} &
\multirow{2}{*}{\textbf{\makecell{Rel.\\AUC}}} &
\multirow{2}{*}{\textbf{\makecell{Avg.\\AUC}}} &
\multirow{2}{*}{\textbf{\makecell{OOD\\AUC}}} \\
\cmidrule(lr){3-3}\cmidrule(lr){4-5}\cmidrule(lr){6-7}\cmidrule(lr){8-9}\cmidrule(lr){10-11}
& &
\textbf{AUC} &
\textbf{AUC} & \textbf{(\%)} &
\textbf{AUC} & \textbf{(\%)} &
\textbf{AUC} & \textbf{(\%)} &
\textbf{AUC} & \textbf{(\%)} &
& & \\
\midrule

\rowcolor{llama!50}
 & checklist  & 0.795 & 0.592 & 74.47\% & 0.526 & 66.15\% & 0.745 & 93.79\% & 0.771 & 97.06\% & 82.87\% & 0.686 & 0.659 \\
\rowcolor{llama!50}
\multirow{-2}{*}{\texttt{\mbox{llama3-1b}}}
 & supervised & 0.927 & 0.731 & 78.84\% & 0.741 & 79.89\% & 0.809 & 87.26\% & 0.875 & 94.39\% & \underline{\textbf{85.09\%}} & 0.816 & \underline{\textbf{0.789}} \\

\rowcolor{llama!100}
 & checklist  & 0.868 & 0.810 & 93.31\% & 0.766 & 88.25\% & 0.889 & 102.39\% & 0.909 & 104.65\% & \underline{\textbf{97.15\%}} & 0.848 & \underline{\textbf{0.843}} \\
\rowcolor{llama!100}
\multirow{-2}{*}{\texttt{\mbox{llama3-3b}}}
 & supervised & 0.940 & 0.747 & 79.43\% & 0.790 & 84.02\% & 0.825 & 87.75\% & 0.944 & 100.41\% & 87.90\% & 0.849 & 0.827 \\

\rowcolor{gemma!50}
 & checklist  & 0.874 & 0.823 & 94.18\% & 0.809 & 92.52\% & 0.893 & 102.17\% & 0.902 & 103.15\% & \underline{\textbf{98.00\%}} & 0.860 & 0.857 \\
\rowcolor{gemma!50}
\multirow{-2}{*}{\texttt{\mbox{gemma3-4b}}}
 & supervised & 0.922 & 0.766 & 83.11\% & 0.826 & 89.62\% & 0.882 & 95.68\% & 0.967 & 104.89\% & 93.32\% & 0.873 & \underline{\textbf{0.861}} \\

\rowcolor{mistral!100}
 & checklist  & 0.888 & 0.801 & 90.27\% & 0.777 & 87.53\% & 0.907 & 102.22\% & 0.930 & 104.81\% & \underline{\textbf{96.21\%}} & 0.861 & \underline{\textbf{0.854}} \\
\rowcolor{mistral!100}
\multirow{-2}{*}{\texttt{\mbox{mistral-7b}}}
 & supervised & 0.945 & 0.706 & 74.70\% & 0.757 & 80.07\% & 0.829 & 87.64\% & 0.893 & 94.47\% & 84.22\% & 0.826 & 0.796 \\

\rowcolor{llama!150}
 & checklist  & 0.870 & 0.708 & 81.38\% & 0.557 & 64.04\% & 0.845 & 97.08\% & 0.909 & 104.48\% & 86.74\% & 0.778 & 0.755 \\
\rowcolor{llama!150}
\multirow{-2}{*}{\texttt{\mbox{llama3-8b}}}
 & supervised & 0.939 & 0.775 & 82.57\% & 0.825 & 87.91\% & 0.868 & 92.53\% & 0.958 & 102.07\% & \underline{\textbf{91.27\%}} & 0.873 & \underline{\textbf{0.857}} \\

\rowcolor{gemma!100}
 & checklist  & 0.890 & 0.738 & 82.88\% & 0.850 & 95.46\% & 0.910 & 102.20\% & 0.949 & 106.61\% & \underline{\textbf{96.79\%}} & 0.867 & 0.862 \\
\rowcolor{gemma!100}
\multirow{-2}{*}{\texttt{\mbox{gemma3-12b}}}
 & supervised & 0.937 & 0.792 & 84.45\% & 0.832 & 88.77\% & 0.886 & 94.51\% & 0.979 & 104.48\% & 93.05\% & 0.885 & \underline{\textbf{0.872}} \\

\rowcolor{mistral!150}
 & checklist  & 0.934 & 0.770 & 82.45\% & 0.794 & 85.04\% & 0.910 & 97.45\% & 0.976 & 104.53\% & \underline{\textbf{92.37\%}} & 0.877 & \underline{\textbf{0.863}} \\
\rowcolor{mistral!150}
\multirow{-2}{*}{\texttt{\mbox{mistral-24b}}}
 & supervised & 0.934 & 0.725 & 77.60\% & 0.739 & 79.17\% & 0.817 & 87.44\% & 0.954 & 102.17\% & 86.60\% & 0.834 & 0.809 \\

\rowcolor{gemma!150}
 & checklist  & 0.900 & 0.735 & 81.61\% & 0.797 & 88.56\% & 0.905 & 100.57\% & 0.946 & 105.02\% & \underline{\textbf{93.94\%}} & 0.857 & 0.846 \\
\rowcolor{gemma!150}
\multirow{-2}{*}{\texttt{\mbox{gemma3-27b}}}
 & supervised & 0.950 & 0.785 & 82.67\% & 0.835 & 87.86\% & 0.867 & 91.21\% & 0.962 & 101.31\% & 90.76\% & 0.880 & \underline{\textbf{0.862}} \\

\midrule
\multicolumn{14}{l}{\textbf{Encoder-only}}\\
\addlinespace[2pt]

\rowcolor{encoder!10}
\texttt{bert}       & supervised & 0.845 & 0.601 & 71.15\% & 0.663 & 78.47\% & 0.737 & 87.21\% & 0.659 & 77.99\% & 78.71\% & 0.701 & 0.665 \\

\rowcolor{encoder!150}
\texttt{roberta}    & supervised & 0.890 & 0.765 & 85.88\% & 0.708 & 79.47\% & 0.808 & 90.70\% & 0.725 & 81.37\% & 84.36\% & 0.779 & 0.751 \\

\rowcolor{encoder!10}
\texttt{deberta}    & supervised & 0.615 & 0.465 & 75.63\% & 0.523 & 84.92\% & 0.537 & 87.21\% & 0.474 & 77.10\% & 81.21\% & 0.523 & 0.500 \\

\rowcolor{encoder!150}
\texttt{modernbert} & supervised & 0.856 & 0.721 & 84.29\% & 0.627 & 73.29\% & 0.755 & 88.25\% & 0.673 & 78.64\% & 81.11\% & 0.726 & 0.694 \\
\bottomrule
\end{tabularx}

\caption{AUC for different models comparing supervised vs.\ \texttt{checklist} approaches, evaluated on multiple test datasets. All models were trained on \textbf{Stormfront} and evaluated both in-domain and out-of-domain . Percent values (\%) report the relative AUC. \textit{Rel.\ AUC} denotes the average relative AUC across all test datasets, \textit{Avg.\ AUC} the mean of the absolute AUC scores, and \textit{OOD AUC} the mean AUC computed by excluding the in-domain evaluation on Stormfront.}
\label{tab:auc_stormfront}
\end{table*}




We evaluate whether the proposed checklist representation improves over direct zero-shot LLM classification (Table~\ref{tab:checklist_vs_zero-shot}).
Additionally, we compare our approach against in-domain supervised fine-tuning under cross-dataset evaluation, training on each of the three datasets that provide a standard train split (\textit{Measuring Hate Speech}, \textit{HateXplain}, and \textit{Stormfront}) and testing on the remaining benchmarks (Tables~\ref{tab:auc_mhs}, \ref{tab:auc_hatexplain} and \ref{tab:auc_stormfront}).
Throughout, we emphasize both absolute AUC and cross-dataset robustness (relative AUC) and the out-of-domain (OOD) AUC averages reported in the tables.

\subsection{Checklist prompting consistently outperforms direct zero-shot classification}
\label{subsec:results_zeroshot}

Table~\ref{tab:checklist_vs_zero-shot} compares direct zero-shot classification with our checklist-based inference across a range of decoder LLMs and evaluation datasets.
For the \texttt{checklist} approach, the reported AUC on dataset \texttt{X} is computed by averaging the performance across training in all datasets \emph{except} \texttt{X}. This design choice mitigates in-domain bias and enables a fairer comparison between zero-shot prediction and the checklist-based framework.

Overall, the checklist formulation yields consistent gains over direct prompting, often by substantial margins.
The improvements are particularly pronounced on datasets that require finer-grained semantic judgments beyond surface toxicity cues.
For instance, on \textit{HateXplain}, gains reach +0.146 AUC for \texttt{llama3-3b}, +0.187 for \texttt{gemma3-4b}, and around +0.18 for multiple mid/large models (e.g., \texttt{mistral-7b}, \texttt{mistral-24b}, \texttt{gemma3-27b}), indicating that decomposing the task into conceptual factors mitigates the brittleness of single-shot hate/neutral decisions.

While improvements are strong across most settings, the smallest model exhibits higher variance: \texttt{llama3-1b} shows small gains on \textit{ETHOS} and \textit{HateCheck} but slight degradations on \textit{HateXplain} and \textit{Stormfront}.
In contrast, for models at 3B parameters and above, the checklist approach improves zero-shot performance across all evaluated datasets. One exceptions is \texttt{Llama3-8b} tested on \textit{Stormfront}, probably because it is quantized to 4 bits, and its overall size is below that of 3B models.

\subsection{Fine-tuning maximizes in-domain performance, but checklist improves cross-dataset robustness}
\label{subsec:results_transfer}

\Cref{tab:auc_mhs,tab:auc_hatexplain,tab:auc_stormfront} compare supervised fine-tuning and checklist-based classification under cross-dataset evaluation.
In each table, models are trained on a single dataset and then evaluated both in-domain and out-of-domain on the remaining benchmarks.
As expected, supervised fine-tuning typically yields the strongest in-domain AUC, reflecting its ability to fit dataset-specific annotation criteria.
However, this advantage does not always translate into robust transfer.

A consistent pattern across training settings is that the checklist approach better preserves performance under domain shift, as reflected by the relative AUC summary.
When training on \textit{Measuring Hate Speech} (~\Cref{tab:auc_mhs}), supervised models achieve very strong in-domain AUC and also higher overall generalization, as measured by OOD AUC. However, this gap closes as the model size increases. Regarding the preservation of performance, we see that the relative AUC with respect to in-domain performance is higher for the checklist in most cases, yielding an average above 100\% for all models.

The same trend holds when training on \textit{HateXplain} (~\Cref{tab:auc_hatexplain}).
Fine-tuned models again maximize in-domain performance, but several exhibit noticeable degradation on other datasets.
Checklist-based models, by contrast, show strong relative transfer, and in multiple cases match or exceed supervised OOD AUC.
Notably, for \texttt{llama3-3b}, \texttt{mistral-7b}, \texttt{mistral-24b} and \texttt{gemma3-27b} the checklist approach yields a higher OOD AUC than supervised fine-tuning.

When training on Stormfront (~\Cref{tab:auc_stormfront}), supervised fine-tuning remains competitive and often superior in absolute terms, but the checklist approach again reduces relative degradation across external datasets.
This is consistent with Stormfront’s narrower domain and older linguistic style: direct supervision can strongly specialize to the training distribution, whereas the checklist representation emphasizes reusable conceptual features that remain meaningful across benchmarks.

Let us now synthesize all results. Across all experiments, supervised models perform better in-domain than our approach, although the gap is smaller as the model size increases. The opposite is the case when measuring the relative AUC, where the checklist tends to perform better. In terms of absolute OOD AUC, we find that supervised models perform better when trained on MHS, but with mixed results when training is HateX and Stormfront. This suggests that the ability of supervised training to generalize is more sensitive to differences in distribution and dataset quality, something to be expected.



\subsection{Encoders remain competitive in-domain but are less robust under domain shift}
\label{subsec:results_encoders}

The encoder-only baselines further contextualize these findings.
While supervised encoder classifiers can achieve strong in-domain performance, their robustness under domain shift is consistently weaker than that of fine-tuned LLMs, and in many cases comparable to or below the checklist-based LLM pipeline.
This pattern suggests that high in-domain scores for encoder models may partially reflect their ability to exploit dataset-specific lexical and statistical regularities, rather than a stable abstraction of the underlying concept of hate speech.
As a result, these models can achieve competitive performance when train and test distributions are closely aligned, yet degrade sharply when confronted with different linguistic styles, target groups, or definitional thresholds.

\section{Analysis and Interpretability}



Beyond aggregate performance, the proposed framework is explicitly designed to support transparent analysis and interpretable decision-making.
In this section, we analyze which diagnostic questions most strongly influence decisions across datasets, model sizes, and model families, and how these influences remain stable under domain shift.
We then inspect the learned decision structure itself and study representative cases where checklist-based predictions diverge from supervised annotations.

\subsection{Checklist question importance across datasets, model sizes, and families}
\label{subsec:analysis_importance}

\begin{figure}
    \centering
    \includegraphics[width=0.9\linewidth]{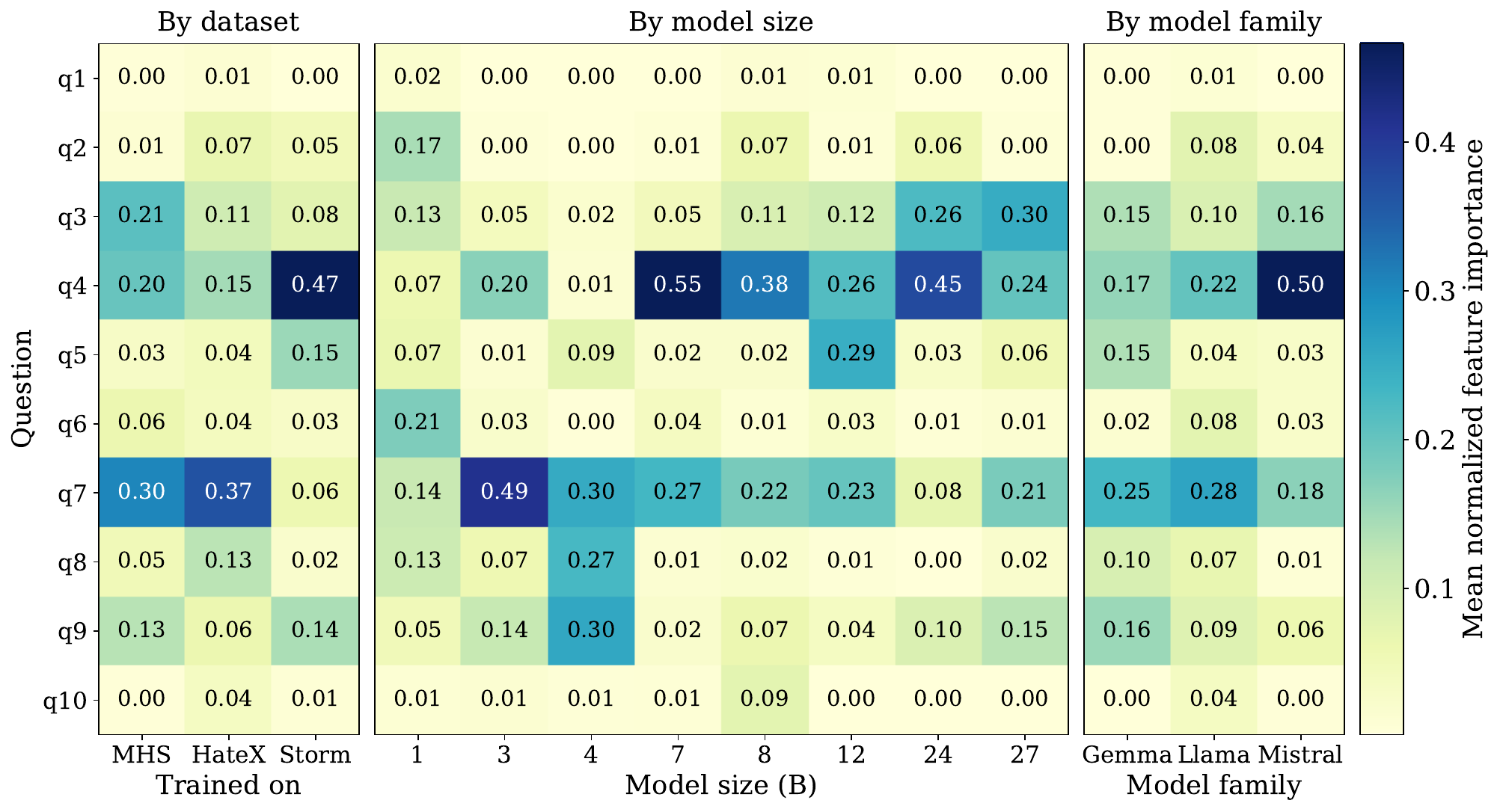}
    \caption{Checklist question importance across training datasets, model sizes (in billions of parameters), and model families.
    Each cell reports the mean normalized importance of a question in the induced decision trees, measured via impurity reduction.}

    \label{fig:feature_importance}
\end{figure}

Figure~\ref{fig:feature_importance} summarizes the mean normalized feature importance of each checklist question when training decision trees on top of their answer, aggregated along three axes: training dataset, model size, and model family.
Feature importance is computed directly from the learned trees and reflects how frequently and how early a given question contributes to splitting the data.

\paragraph{Consistent dominance of a small subset of factors.}
The most influential questions are, in order, \texttt{q4} (dehumanization or vilification), \texttt{q7} (threats or wishes of harm), \texttt{q3} (explicit slur usage), \texttt{q9} (speaker endorsement), and \texttt{q8} (incitement or endorsement of violence).
These questions account for the majority of the splitting power in the learned trees, regardless of the dataset used for training, the size of the underlying LLM, or the model family.

Datasets may differ in annotation guidelines and severity thresholds, yet the same core factors repeatedly emerge as decisive.
This suggests that these questions capture the most discriminative conceptual dimensions of hate speech, and that they align well with the implicit criteria encoded in diverse annotation regimes.

\paragraph{Low importance of definitional gate conditions.}
By contrast, questions \texttt{q1} (protected target identification), \texttt{q2} (derogatory or hostile tone), and \texttt{q10} (perceived harm to the target) exhibit consistently low importance across all settings.
At first glance, this might appear to contradict the theoretical motivation of the checklist, as these questions encode core definitional or impact-oriented conditions.

However, this behavior admits a natural interpretation.
Highly discriminative factors such as dehumanization (\texttt{q4}) or threats of harm (\texttt{q7}) almost never occur in isolation: they presuppose the existence of a target and an attacking stance.
As a result, questions like \texttt{q1} may be strongly correlated with downstream severity signals and may be rendered redundant from the perspective of tree-based splitting.
In this sense, \texttt{q1} can be viewed as a semantic prerequisite or subset of higher-level factors rather than an independent discriminator.

\paragraph{Effect of model size on slur sensitivity.}
A clear trend emerges when stratifying feature importance by model size.
Larger models assign substantially higher importance to \texttt{q3} (explicit slur usage) compared to smaller ones.
This effect is consistent across datasets and model families.

A plausible explanation is that larger LLMs possess broader lexical coverage and richer exposure to niche or emerging online slurs, enabling more reliable detection of explicit identity-based epithets.
Smaller models may fail to recognize less common or context-dependent slurs, reducing the usefulness of \texttt{q3} as a splitting feature in those settings.
As model capacity increases, slur detection becomes a more reliable and informative signal, and the decision trees adapt accordingly.

\subsection{Case-level analysis of checklist and supervised disagreements}
\label{subsec:analysis_disagreements}

\begin{figure}
    \centering
    \includegraphics[width=0.9\linewidth]{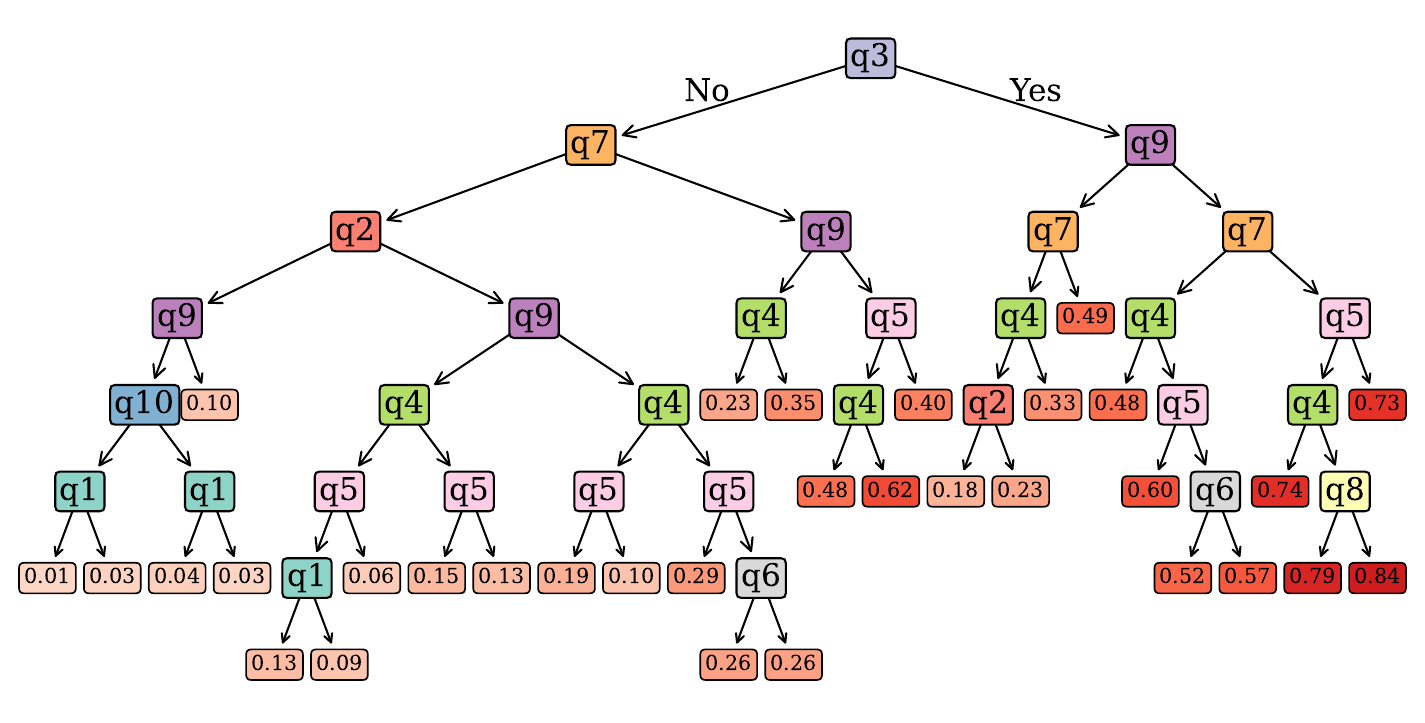}
    \caption{Decision tree learned on top of the checklist representations produced by \texttt{mistral-24b}, trained on the \textit{Measuring Hate Speech} dataset. Each internal node corresponds to a checklist question, and each root-to-leaf path encodes a conjunction of diagnostic decisions that the model treats as sufficient for classification. Leaf nodes report the proportion of samples labeled as hate speech among the training instances that reach that leaf.}
    \label{fig:decision_tree}
\end{figure}

\begin{table*}[!ht]
\scriptsize
\centering
\setlength{\tabcolsep}{3pt}
\renewcommand{\arraystretch}{1.15}

\begin{tabularx}{\textwidth}{
>{\RaggedRight\arraybackslash}p{0.3\textwidth}
>{\RaggedRight\arraybackslash}p{0.2\textwidth}
>{\RaggedRight\arraybackslash}p{0.35\textwidth}
>{\centering\arraybackslash}p{0.05\textwidth}
>{\centering\arraybackslash}p{0.05\textwidth}
}
\toprule
\textbf{Input text} &
\textbf{Decision path (checklist)} &
\textbf{Model rationales (per decision)} &
\textbf{Checklist pred.} &
\textbf{Gold label} \\
\midrule

\textit{This was one of the first instances that I learned that women ain't shit} &
$q_3=\text{Yes} \rightarrow q_9=\text{Yes} \rightarrow q_7=\text{No} \rightarrow q_4=\text{Yes} \rightarrow q_5=\text{No}$ &
\begin{minipage}[t]{\linewidth}
\begin{itemize}[leftmargin=*,nosep, label={}]
\item \textit{q3:} The phrase ``ain't shit'' is an epithet.
\item \textit{q9:} Speaker expresses own belief (not quoting/joking).
\item \textit{q7:} No threat of harm.
\item \textit{q4:} ``ain't shit'' is dehumanizing.
\item \textit{q5:} Not blaming for societal problems; expresses worthlessness.
\end{itemize}
\end{minipage}
&
1 (0.60) &
1
\\

\addlinespace

\textit{Niggas be dirty as shit in real life on my kidss} &
$q_3=\text{Yes} \rightarrow q_9=\text{Yes} \rightarrow q_7=\text{No} \rightarrow q_4=\text{Yes} \rightarrow q_5=\text{No}$ &
\begin{minipage}[t]{\linewidth}
\begin{itemize}[leftmargin=*,nosep, label={}]
\item \textit{q3:} ``niggas'' is a racial slur.
\item \textit{q9:} Used derogatorily; speaker endorses the insult.
\item \textit{q7:} No threat of harm.
\item \textit{q4:} Slur + ``dirty'' are dehumanizing.
\item \textit{q5:} Not scapegoating; slur/stereotype without societal blame.
\end{itemize}
\end{minipage}
&
1 (0.60) &
0
\\

\addlinespace

\textit{How a pussy ass lame ass dirty ass fuck nigga kill a real ass nigga dawg} &
$q_3=\text{Yes} \rightarrow q_9=\text{No} \rightarrow q_7=\text{Yes}$ &
\begin{minipage}[t]{\linewidth}
\begin{itemize}[leftmargin=*,nosep, label={}]
\item \textit{q3:} Contains explicit slurs.
\item \textit{q9:} Interpreted as quotation (e.g., lyrics), not endorsement.
\item \textit{q7:} Expresses a desire for harm.
\end{itemize}
\end{minipage}
&
0 (0.51) &
1
\\

\addlinespace

\textit{Why don't you take the key and fuck her you loser piece of shit?} &
$q_3=\text{No} \rightarrow q_7=\text{Yes} \rightarrow q_9=\text{Yes} \rightarrow q_5=\text{No} \rightarrow q_4=\text{No}$ &
\begin{minipage}[t]{\linewidth}
\begin{itemize}[leftmargin=*,nosep, label={}]
\item \textit{q3:} No slur or epithet.
\item \textit{q7:} Threat of sexual violence.
\item \textit{q9:} Clearly insulting/demeaning intent.
\item \textit{q5:} No identity-based scapegoating.
\item \textit{q4:} Not dehumanizing a group; targets an individual.
\end{itemize}
\end{minipage}
&
0 (0.52) &
0
\\

\addlinespace

\textit{Well I mean its justified when you lunatics screech anti -police rhetoric . Tell me if armed hardcore conservative Muslims were to trash your "pride parade" , who would you wish for to be standing by to dissolve the situation.} &
$q_3=\text{Yes} \rightarrow q_9=\text{Yes} \rightarrow q_7=\text{No} \rightarrow q_4=\text{Yes} \rightarrow q_5=\text{Yes} \rightarrow q_6=\text{Yes}$ &
\begin{minipage}[t]{\linewidth}
\begin{itemize}[leftmargin=*,nosep, label={}]
\item \textit{q3:} Epithets / negative stereotypes (e.g., ``lunatics'').
\item \textit{q9:} Speaker endorses hostile framing (not quoting).
\item \textit{q7:} No explicit threat.
\item \textit{q4:} Dehumanization + demonization as a threat.
\item \textit{q5:} Identity-based accusations of wrongdoing.
\item \textit{q6:} Implies exclusionary treatment / removal.
\end{itemize}
\end{minipage}
&
1 (0.57) &
0
\\

\bottomrule
\end{tabularx}

\caption{Selected samples where checklist-based tree predictions and supervised LLM predictions differ. Each row shows the input, the induced decision path, and per-decision rationales. We took samples from the \emph{Measuring Hate Speech} dataset and used predictions from the \texttt{mistral-24b} backbone model.}
\label{tab:checklist_contradictions}
\end{table*}

To ground the interpretability analysis, we analyze the decision tree learned on top of the checklist responses produced by \texttt{mistral-24b}, trained using \textit{Measuring Hate Speech} as the source dataset as it corresponds to the setting in which checklist-based and supervised models achieve their highest and most closely matched OOD performance (see Figure~\ref{fig:decision_tree}).

Building on this representation, Table~\ref{tab:checklist_contradictions} presents a small set of examples where checklist-based predictions and supervised LLM predictions disagree.
These cases illustrate distinct and recurrent sources of error in hate speech datasets and highlight how the checklist behaves under annotation noise, contextual ambiguity, and borderline definitions.

The first example corresponds to a straightforward instance of identity-based derogation.
Here, the prediction of the checklist aligns with the intuitive interpretation of the text as hate speech, activating slur-related predicates, endorsement, and dehumanization.
The supervised prediction is negative, suggesting either an overly permissive annotation or inconsistent treatment of misogynistic language.

The second example is more revealing.
Despite the presence of an explicit racial slur and a demeaning stereotype, the gold label is negative.
This appears to be a clear misannotation rather than a genuine conceptual disagreement.
Importantly, the checklist does not adapt to this noise: it follows the same decision path as in similar, correctly labeled cases.
This highlights a key property of the framework: although the downstream decision tree is learned, the diagnostic space itself is constrained by explicit conceptual questions answered in an unsupervised manner, preventing the system from overfitting to isolated annotation errors.

The third example involves a song lyric (\textit{"Pussy Nigga"} by \textit{Woop} - 2014).
Although it contains slurs and references to violence, the checklist correctly identifies the lack of endorsement by the speaker ($q_9=\text{No}$), leading to a negative prediction.
The positive gold label suggests that contextual quotation or artistic expression is not consistently taken into account in the dataset.
This case illustrates the practical value of explicitly modeling contextual endorsement: the checklist makes transparent why such examples are rejected as hate speech, even when surface-level cues might mislead a supervised model.

The fourth example consists of a severe personal insult involving slurs, but directed at an individual rather than a protected group.
Consequently, the checklist rejects it based on the absence of group-level targeting and dehumanization.
This aligns with most formal definitions of hate speech, which distinguish between abusive language and identity-based hostility.
The positive annotation again points to a broader tendency in some datasets to collapse hate speech and generic abuse into a single category, a distinction the checklist is designed to preserve.

Finally, the last example is the most subtle.
The text relies on an implicit framing that invokes stereotypes about Muslims and LGBTQ+ communities without explicitly stating a threat or a heavy slur attack.
Correctly interpreting hateful intent requires integrating multiple cues in the message.
Despite this difficulty, the checklist activates a consistent chain of predicates (derogation, endorsement, dehumanization, identity-based accusation, and implied exclusion), leading to a positive prediction that aligns with a careful human reading.
This case demonstrates that the framework is not limited to detecting explicit hate, but can also capture indirect and context-dependent forms of hostility.

Together, these examples show that the framework remains notably robust to noisy or inconsistent annotations.
Rather than bending to mislabeled samples, the checklist enforces a stable conceptual definition of hate speech and exposes disagreements explicitly.
This makes errors interpretable, debuggable, and, crucially, attributable either to annotation artifacts or to well-defined conceptual assumptions, rather than to opaque model behavior.

\section{Conclusion}


We proposed a diagnostic, factorized framework that decomposes hate speech into minimal, objective, interpretable predicates, evaluated independently by an LLM and aggregated through a simple decision tree.
Across datasets, this approach consistently improves cross-dataset robustness, often approaching or matching supervised baselines while providing explicit and inspectable reasoning paths.

Our results indicate that encoder-based hate speech detection models achieve strong in-domain performance, but degrade substantially under domain shift.
This degradation is consistently larger than that observed for causal LLMs and checklist-based pipelines, as reflected by lower relative AUC and out-of-domain AUC values.
While checklist-based methods do not always outperform supervised fine-tuning in absolute OOD performance, they consistently surpass zero-shot LLM classification and exhibit more stable performance across datasets, indicating improved robustness when moving away from the training distribution.

The main limitations of our method are the higher inference cost--requiring multiple LLM calls per input--and, in some settings, slightly lower absolute out-of-domain performance compared to supervised fine-tuning.
These drawbacks are partially offset by the inherent interpretability of the model, which exposes how individual conceptual factors contribute to each decision.

Another key advantage of the proposed framework is its robustness to annotation noise.
Rather than adapting to mislabeled or inconsistent samples, the checklist enforces a stable conceptual structure, making disagreements explicit and attributable to either annotation artifacts or definitional ambiguity.
This suggests its potential use as a diagnostic tool for identifying and mitigating dataset bias.

Finally, although evaluated on hate speech detection, the framework is model-agnostic and task-independent.
Any classification problem that can be expressed as a set of interpretable semantic factors may benefit from this approach, making it a general strategy for building robust and explainable NLP systems.

\bibliographystyle{ieeetr}
\bibliography{references}

\clearpage
\appendix

\section{Checklist questions and rationales.}
\label{sec:questions_and_rationales}

\newcolumntype{Y}{>{\RaggedRight\arraybackslash}X}

\begin{table*}[!ht]
\footnotesize
\centering
\setlength{\tabcolsep}{5pt}
\renewcommand{\arraystretch}{1.25}
\begin{tabularx}{\textwidth}{
p{0.1\textwidth} p{0.28\textwidth} p{0.55\textwidth}
}
\toprule
 & \textbf{Question} & \textbf{Rationale} \\
\midrule

Q1 (Protected Group Referenced) &
Does the text target—explicitly or implicitly—any person or group because of an inherent protected characteristic
(race, ethnicity, nationality, religion, gender, sexual orientation, disability, age, etc.)? &

This question operationalizes the protected characteristic factor.
It forces the evaluator to identify whether hostility is tied to identity.
By listing typical protected traits in the question, we remind the evaluator what counts.
This addresses false positives where harsh language isn't actually identity-based.
Political belief, occupation, fandom, ideology, etc are **not** protected categories.
\\

Q2 (Negative or Derogatory Description) &
Is the text derogatory or insulting toward a group or person because of a protected trait?
Does it use slurs, name-calling, or language that conveys hatred, contempt, or inferiority? &

This checks for "problematic content", i.e. negative tone or intent.
If someone references a group but says nothing negative, it's not hate. This question captures generic insults or negative adjectives that indicate contempt or hatred.
\\

Q3 (Use of Hateful Slurs or Stereotypes) &
Does the text use any explicit slurs, epithets, or negative stereotypes associated with a group? (Slurs are strong indicators of hate speech) &

Slurs are strong indicators of hate speech. Stereotypes might be statements like "They're all terrorists”, suggesting a predefined hateful generalization. Slurs are often considered inherently hateful (with few exceptions for reappropriation context).
\\

Q4 (Dehumanization or Vilification) &
Does the text dehumanize some group or compare them to subhuman entities or demonize them as a threat? &

Dehumanization is highlighted in virtually every serious hate speech framework.
It's a hallmark of content that can incite violence because it psychologically conditions people to view the target as not human.
It might not always use obvious slurs, so it deserves a dedicated check.
\\

Q5 (Blaming or Scapegoating) &
Does the text blame some group for societal problems or accuse them of wrongdoing based solely on their identity? &

This question ensures we catch statements that are not just name-calling but spread harmful stereotypes or conspiracy theories. Many legal definitions implicitly include this (incitement to hostility often comes by painting the group as bad actors).
\\

Q6 (Calls for Exclusion or Discrimination) &
Does the text advocate or encourage excluding some group or denying them equal rights? &

This addresses content aimed at rallying others to take negative action (social, legal, economic) against a group. So we want to capture those non-violent but harmful calls.
\\

Q7 (Threats or Wishes of Harm) &
Does the text express a desire for harm or violence toward some group (even implicitly), or threaten them? &

This isolates explicit violent intent. If someone says "I wish they would all die" or "they should be shot" that's a direct expression of violent hatred.
Violence is a qualitative leap in severity. This question ensures the evaluator checks for words like "kill, shoot, exterminate, hang," etc.
\\

Q8 (Direct Incitement of Violence) &
Is the text directly calling on others to commit violence or attacks against some group imminently? &

This specifically checks for incitement (though we won't always have a timeframe, "near term" is to hint at imminence). Here is the speaker trying to rally others to act. It's legally a key threshold (speech that solicits others to commit violence is generally unprotected worldwide).
\\

Q9 (Context and Intent Assessment) &
Taking the text in context, does it appear the speaker is serious and endorsing the hateful message, rather than quoting, reporting, or joking? &

This is our context/intent filter. We included this to catch false positives. This question essentially asks the evaluator to step back: "Given how this text is presented, do we interpret it as actually advocating hate?"
\\

Q10 (Likely Effect on Targeted Group) &
Would a person from the targeted group reasonably feel attacked, threatened, or demeaned by this content purely because of their group identity? &

This question thus helps differentiate broad criticism from hateful attack. It also might surface subtle hate: sometimes someone uses coded language or mild tone but if you belong to the group, you know it's hostile.
\\

\bottomrule
\end{tabularx}
\caption{Pairs of checklist questions and their corresponding rationales.
The rationales shown here are lightly adapted for clarity and brevity.
The original formulations used in the experiments are available in the implementation.}

\label{tab:checklist_prompts}

\end{table*}


\begin{table*}[!ht]
\scriptsize
\centering
\setlength{\tabcolsep}{3pt}
\renewcommand{\arraystretch}{1.15}
\begin{tabularx}{\textwidth}{
>{\RaggedRight\arraybackslash}p{0.10\textwidth}
*{10}{p{0.075\textwidth}}
}
\toprule
\textbf{Sample} &
\textbf{Q1} & \textbf{Q2} & \textbf{Q3} & \textbf{Q4} & \textbf{Q5} &
\textbf{Q6} & \textbf{Q7} & \textbf{Q8} & \textbf{Q9} & \textbf{Q10} \\
\midrule

\textit{All [slur] should be shot. We need to rid the world of those animals.} &
\cellcolor{yes}It targets a group by slur, e.g. an ethnic slur. &
\cellcolor{yes}It is very derogatory: calls them ``animals''. &
\cellcolor{yes}It uses an explicit slur. &
\cellcolor{yes}Dehumanization: calls them animals; ``rid the world''. &
\cellcolor{yes}Implies they are a plague to ``rid''. &
\cellcolor{yes}Advocates removing them from the world. &
\cellcolor{yes}Explicit wish of violence: ``should be shot''. &
\cellcolor{yes}Calls others to action (``we'' / implicit urging). &
\cellcolor{yes}Clearly serious and hateful; no irony. &
\cellcolor{yes}A targeted person would feel terrified and attacked.
\\

\textit{I hate Muslims. They are disgusting and backward.} &
\cellcolor{yes}It targets ``Muslims'', a religious group. &
\cellcolor{yes}It has hateful tone. &
\cellcolor{no}No explicit slur (common words only). &
\cellcolor{yes}Vilifying language (``disgusting'', ``backward''). &
\cellcolor{no}Insulting, but not scapegoating for a broader problem. &
\cellcolor{no}No direct call for exclusion/discrimination. &
\cellcolor{no}No threat or wish of harm. &
\cellcolor{no}No incitement of others. &
\cellcolor{yes}Speaker endorses the hate; not quoting. &
\cellcolor{yes}A Muslim reader would feel attacked for identity.
\\

\textit{The word ``<slur>'' is a derogatory term used to insult <group>.} &
\cellcolor{yes}It references a protected group. &
\cellcolor{no}The text itself isn't insulting the group; it's explaining a term. &
\cellcolor{yes}The slur appears in the text. &
\cellcolor{no}No dehumanization; academic tone. &
\cellcolor{no}No blame not trait ascribed; about the word. &
\cellcolor{no}No call for discrimination; just explanation. &
\cellcolor{no}No threat. &
\cellcolor{no}No incitement. &
\cellcolor{no}Intent is descriptive and condemning the slur. &
\cellcolor{no}Not an attack; describes usage (though slur is present).
\\

\textit{Ugh, gamers are such basement-dwelling losers.} &
\cellcolor{no}Gamers aren't a protected trait. &
\cellcolor{yes}Yes it's derogatory. &
- &
- &
- &
- &
\cellcolor{no}No threat of harm. &
- &
\cellcolor{yes}Intent is insulting (even if not protected-target). &
\cellcolor{yes}A gamer might feel insulted.
\\

\textit{I think <immigrant group> tend to commit more crimes. It's just facts.} &
\cellcolor{yes}It targets an immigrant group. &
\cellcolor{yes}Negative claim (group defamation) even without slurs. &
\cellcolor{no}No explicit slur; just a negative claim. &
\cellcolor{yes}Vilifies them as criminals and dangerous. &
\cellcolor{yes}Scapegoats the group for crime issues. &
\cellcolor{no}No explicit call for exclusion nor discrimination. &
\cellcolor{no}No threat or desire of harm expressed. &
\cellcolor{no}No direct incitement. &
\cellcolor{yes}Speaker endorses the claim as ``facts''. &
\cellcolor{yes}Group would feel labeled as criminals (xenophobic).
\\

\textit{We need a final solution for the <group> problem.} &
\cellcolor{yes}It references a protected group. &
\cellcolor{yes}Derogatory framing: group as a ``problem''. &
\cellcolor{no}No explicit slur; euphemistic phrasing. &
\cellcolor{yes}Treats group as pestilence to be ``solved'' (dehumanizing). &
\cellcolor{yes}Frames group as responsible for an issue; needs removal. &
\cellcolor{yes}Call for elimination and exclusion (``final solution''). &
\cellcolor{yes}Thinly veiled call for mass harm. &
\cellcolor{yes}Incites action against the group (implied violent). &
\cellcolor{yes}Clearly advocating hate. &
\cellcolor{yes}Targeted person would feel extremely threatened.
\\

\textit{Ugh, men are trash. Never trust a man.} &
\cellcolor{yes}It targets men (protected by gender). &
\cellcolor{yes}Derogatory (``trash''). &
\cellcolor{no}``Trash'' is an insult, not a slur or epithet. &
\cellcolor{yes}Dehumanizing (calling people garbage). &
\cellcolor{yes}Generalizes a negative trait to the group (``trash''). &
\cellcolor{yes}Encourages exclusionary treatment (``never trust''). &
\cellcolor{no}No threat of harm. &
\cellcolor{no}No call for violence. &
\cellcolor{yes}Likely earnest endorsement (not satire). &
\cellcolor{yes}Many men would feel attacked and sexist.
\\

\textit{I don't agree with <religion>'s teachings. I think that religion has done harm in the world.} &
\cellcolor{yes}Religion is tied to believers (borderline). &
\cellcolor{no}Critical opinion on an institution and belief system. &
\cellcolor{no}No slurs or epithets; disagreement only. &
\cellcolor{no}No dehumanization; ideology critique. &
\cellcolor{no}Blames an ideology, not scapegoating a people. &
\cellcolor{no}No call for discrimination; just disagreement. &
\cellcolor{no}No threats. &
\cellcolor{no}No incitement. &
\cellcolor{no}Sincere criticism, not hate endorsement. &
\cellcolor{no}Subjective, but framed as critique rather than attack.
\\

\textit{There's a reason [derogatory code word] hang from trees.} &
\cellcolor{yes}If code word is recognizable, group is implied. &
\cellcolor{yes}Extremely derogatory (implies deserved lynching). &
\cellcolor{no}No explicit slur; insult via implication and metaphor. &
\cellcolor{yes}Dehumanizing + violent imagery (objects to destroy). &
\cellcolor{no}Justifies violence more than scapegoats for an issue. &
\cellcolor{no}Not exclusion; it references murder and violence. &
\cellcolor{yes}Implies they should be killed. &
\cellcolor{yes}Incitement by condoning and glorifying lynching. &
\cellcolor{yes}Intent clearly hateful, not quoting. &
\cellcolor{yes}A targeted person would feel terrified.
\\

\textit{We should kill every last one of those traitors who voted for that policy.} &
\cellcolor{no}Political affiliation is not a protected category. &
\cellcolor{yes}Derogatory (``traitors''). &
- &
- &
- &
- &
\cellcolor{yes}Death threat broadly. &
- &
\cellcolor{yes}Serious and violent intent. &
\cellcolor{yes}Targeted group would feel threatened.
\\

\bottomrule
\end{tabularx}
\caption{Shared few-shot examples used across checklist questions. Each cell contains a condensed expected rationale and is shaded according to the corresponding binary answer (Yes-green/No-red). Question definitions are provided in Table~\ref{tab:checklist_prompts}.}
\label{tab:fewshot_matrix}
\end{table*}

\clearpage
\end{document}